\documentclass[lettersize,journal]{IEEEtran}
\usepackage{amsmath,amsfonts}
\makeatletter
\newif\if@restonecol
\makeatother

\usepackage[linesnumbered,ruled,vlined]{algorithm2e}%[ruled,vlined]{
\usepackage{algpseudocode}
\usepackage{amsmath}
\usepackage{array}
\usepackage[caption=false,font=normalsize,labelfont=sf,textfont=sf]{subfig}
\usepackage{textcomp}
\usepackage{stfloats}
\usepackage{url}
\usepackage{verbatim}
\usepackage{graphicx}
\usepackage{threeparttable} % 用于表格注释
\usepackage{booktabs}
\usepackage{multicol}
\usepackage{makecell} % 引入 makecell 包用于单元格内的换行
\renewcommand{\arraystretch}{1.2} % 设置行高为正常行高的×倍
\usepackage{multirow}
\usepackage{tabularx}
\usepackage{pifont}
\newcommand{\cmark}{\ding{51}} % check
\newcommand{\xmark}{\ding{55}} % cross

\usepackage{cite}
\usepackage{bm}
\usepackage{xfrac}
\usepackage{hyperref}
\hypersetup{
	colorlinks=true, % 启用彩色链接
	linkcolor=blue,  % 设置链接颜色为蓝色
	citecolor=blue,  % 设置引用颜色为蓝色
	urlcolor=blue,    % 设置URL颜色为红色，直接在这里修改url颜色
	pdfborder={0 0 0} % 设置 PDF 边框为透明 (0 0 0 表示没有边框)
}
\begin{document}
	
\bstctlcite{BSTcontrol}

%%%%%%%%% TITLE
\title{EIVE: End-to-End Instance-Specific Visual Explanations for Detection Transformers}

%%%%%%%%% AUTHORS
\author{
	Jianlin~Xiang,
	Yanshan~Li,
	and~Linhui~Dai
	\thanks{This work was partially supported by the National Natural Science Foundation of China (No.~62471317), the Natural Science Foundation of Shenzhen (No.~JCYJ20240813141331042), and the Guangdong Provincial Key Laboratory of Intelligent Information Processing (Grant~2023B1212060076). \textit{(Corresponding author: Linhui Dai.)}}
	\thanks{Jianlin Xiang, Yanshan Li, and Linhui Dai are with the Institute of Intelligent Information Processing, Shenzhen University; the Guangdong Provincial Key Laboratory of Intelligent Information Processing, Shenzhen University; and the Shenzhen Key Laboratory of Modern Communications and Information Processing, Shenzhen University, Shenzhen, China 
		(e-mail: xiangjianlin2023@email.szu.edu.cn; lys@szu.edu.cn; dailinhui@szu.edu.cn).}
}

\markboth{Journal of \LaTeX\ Class Files,~Vol.~14, No.~8, August~2021}%
{Shell \MakeLowercase{\textit{et al.}}: A Sample Article Using IEEEtran.cls for IEEE Journals}

\IEEEpubid{0000--0000/00\$00.00~````\copyright~2021 IEEE}

\maketitle

\begin{abstract}
	Visual explainability for object detection remains challenging due to the multi-instance nature of detection. Existing approaches predominantly adopt post-hoc paradigms, such as gradient-based or perturbation-based explanation methods, to interpret pretrained detectors. However, these methods require additional gradient computation or repeated model inference, resulting in limited efficiency. To address this issue, we propose an End-to-end Instance-specific Visual Explanation framework (EIVE) that directly generates instance-level saliency maps following the forward pass of Detection Transformer (DETR)-like models. Specifically, we reformulate the cross-attention mechanism in the decoder as an instance-level feature attribution pathway, so that the cross-attention of each object query corresponds to the visual attribution of its predicted instance. Based on this formulation, we design a cross-layer hybrid consensus fusion (CLHCF) module to aggregate cross-attention signals across decoder layers, producing stable and compact explanations. The explanation process of EIVE requires neither gradient computation nor input perturbation, yielding high computational efficiency, and applies to single- and multi-scale DETR-like object detectors. Finally, we present an attention-aware joint training strategy (AAJTS) as a training-oriented application, which imposes spatial constraints on cross-attention patterns to encourage stable and concentrated attribution representations, thereby improving both interpretability and detection performance. Experiments on MS COCO 2017, ExDark, and Cityscapes demonstrate that EIVE produces high-quality instance-level saliency maps and achieves performance comparable to, or better than, state-of-the-art post-hoc methods across standard metrics, while substantially improving explanation efficiency. Code is available at https://github.com/xjlDestiny/EIVE.git.
\end{abstract}


%     \link{Code}{https://github.com/xjlDestiny/HiProto.git}
%     % \link{Datasets}{https://aaai.org/example/datasets}
%     % \link{Extended version}{https://aaai.org/example/extended-version}

\section{Introduction}

\begin{figure}[!t]
	\centering
	\includegraphics[width=\columnwidth]{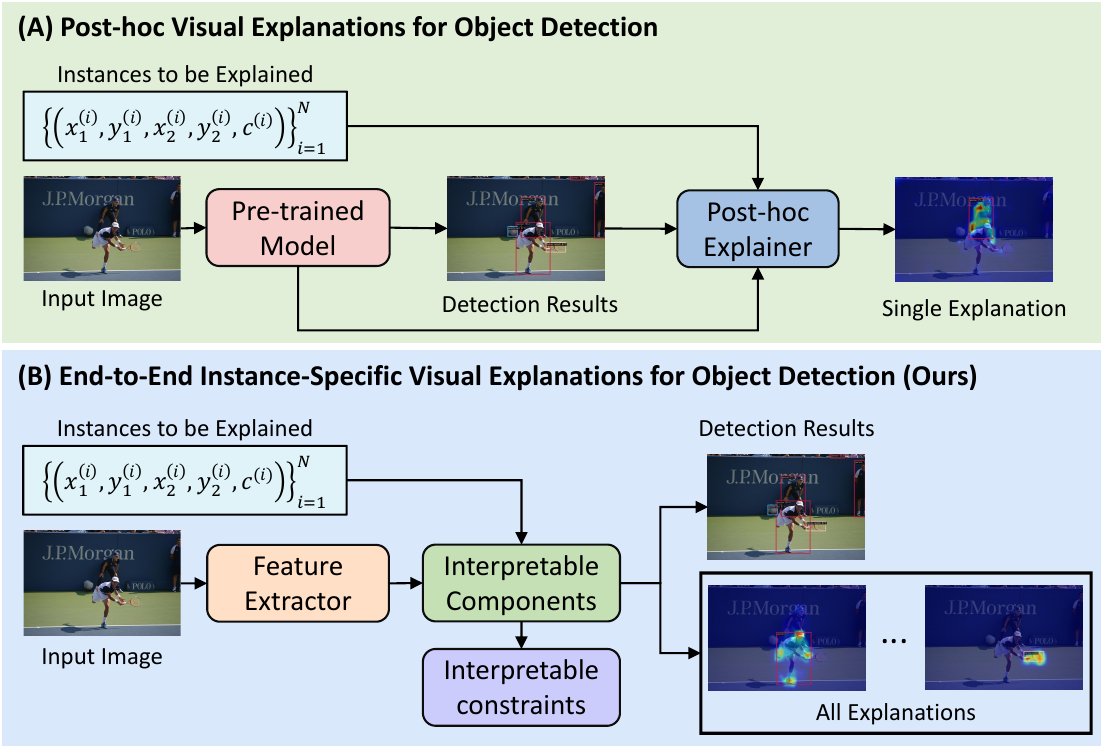}
	\caption{Comparison of post-hoc and end-to-end instance-specific visual explanations for object detection. Our method produces detection results and then derives instance-specific explanations from the forward-pass outputs within a unified framework.}
	\label{fig:Abstract-diagram}
\end{figure}

Deep neural networks~\cite{ref_Residual, ref_Shufflenet, ref_Mobilenetv2, ref_SBSNet} have achieved remarkable success, yet their highly nonlinear decision-making processes remain difficult to interpret~\cite{ref_PRFA, ref_BBAAOB}. This lack of transparency limits model debugging, reliability assessment, and deployment in safety-critical scenarios~\cite{ref_PODAD, ref_Safe}. To address this issue, explainability methods aim to reveal the discriminative evidence underlying model predictions, thereby helping researchers understand model behavior and providing interpretable support for decision-making.

Most existing studies focus on post-hoc visual explanation, where pretrained models are interpreted after training without modifying the original architecture or learning procedure. This paradigm has been extensively studied in image classification and has formed a relatively mature framework, including CAM-based methods~\cite{ref_CAM, ref_LayerCAM, ref_Grad-CAM, ref_Grad-CAM++, ref_Score-CAM, ref_SSGrad-CAM++, ref_Finer-CAM, ref_BI-CAM, ref_CR-CAM, ref_GT-CAM}, input perturbation-based methods~\cite{ref_LIME, ref_RISE, fong2017interpretable, fong2019understanding, ref_IGOS, ref_IGOS++}, and SHAP-based methods~\cite{ref_SHAP, ref_DeepSHAP}. By analyzing the sensitivity of predictions to inputs or intermediate features, these approaches generate saliency maps that provide intuitive explanations for classification decisions. However, extending such methods directly to object detection~\cite{ref_RCNN, ref_FastRCNN, ref_FasterRCNN, ref_Mask_RCNN, ref_YOLO, ref_RetinaNet, ref_FCOS, ref_DETR} remains challenging. Unlike classification, object detection must explain both what is detected and where it is located, often in the presence of multiple spatially independent or overlapping instances. As a result, explanations for detection models must preserve both spatial localization and instance-level discriminability, which substantially increases the difficulty of explainability modeling.

\IEEEpubidadjcol
Recent studies have attempted to adapt post-hoc explanation methods to object detection by generating instance-level saliency maps for target instances and evaluating them through protocols such as insertion-deletion. However, these methods usually rely on external explanation operators to produce additional explanatory signals, such as additional gradient computation~\cite{ref_ODAM} or repeated input perturbations~\cite{ref_D-RISE, ref_D-HSIC, ref_VPS}, resulting in considerable time and memory overhead. More importantly, since these saliency maps are generated mainly from external procedures rather than the detector’s internal reasoning signals, they are less consistent with the actual prediction mechanism and difficult to effectively support downstream applications or serve as meaningful constraints for detection optimization. In contrast, an end-to-end visual explanation framework unifies prediction and explanation within the same inference process, thereby enabling more efficient, stable, and semantically consistent instance-level interpretations, while also making the resulting explanations more amenable to downstream use and to serving as effective supervisory signals for detection optimization. Therefore, we seek to explore whether object detection models can generate instance-specific saliency maps directly from forward-pass outputs in an end-to-end manner, without requiring gradients or repeated perturbations.

To enable end-to-end explainability, whether a detector explicitly models object instances in its architecture becomes critical. Detection Transformer (DETR)~\cite{ref_DETR} and its variants~\cite{ref_deformable, ref_conditional, ref_dab, ref_dino} provide a particularly suitable foundation. By formulating detection as set prediction with learnable object queries, DETR-like models establish an intrinsic one-to-one correspondence between queries and object instances while capturing global contextual relationships through attention. In particular, each object query interacts with image features through decoder cross-attention, selectively aggregating instance-relevant information from spatial feature maps. This mechanism naturally defines an instance-level feature attribution pathway and creates an internal correspondence between predictions and input regions. Existing studies on Transformer interpretability~\cite{ref_Transformer, jain2019attention, ref_Attention-Rollout, ref_Attention-Visualization, Chefer_2021_ICCV, ref_AttCAT, ref_TokenTM, ref_FaithfulnessViT, ref_DETD} further suggest that such architectures contain exploitable explanatory signals.

Motivated by these observations, we propose End-to-end Instance-specific Visual Explanations (EIVE), an end-to-end visual explanation method based on the cross-attention mechanism. Specifically, we model the cross-attention in the DETR decoder as an instance-level feature attribution pathway, enabling the model to derive an instance-specific saliency map for each detected object from forward-pass cross-attention signals, without requiring any additional gradient computation or input perturbation.

In the DETR architecture, decoder cross-attention is driven by object queries and naturally performs instance-level feature selection, where different queries aggregate shared encoder features to form discriminative representations for different instances. As decoder layers are stacked, both query representations and their cross-attention distributions evolve, reflecting the model’s changing focus on instance-relevant regions. Building on this property, we further propose a cross-layer hybrid consensus fusion (CLHCF) module. For a target instance, CLHCF selects the most relevant query from the last decoder layer and tracks it across selected layers, based on which layer-wise instance-level saliency maps are reconstructed in the original feature space. To obtain a stable final explanation, we further introduce adaptive layer weighting and a hybrid consensus fusion strategy, where the arithmetic branch preserves complementary support across layers and the geometric branch emphasizes cross-layer consistent regions. Since the entire process relies only on prediction outputs and cross-attention signals generated during the forward pass, it avoids additional gradient computation and repeated perturbation, thereby reducing computational cost while preserving semantic consistency.

Furthermore, as a training-oriented application, we present an attention-aware joint training strategy (AAJTS). By imposing spatial constraints on cross-attention patterns, AAJTS encourages the model to learn more stable and concentrated instance-level attribution representations, thereby improving saliency-map faithfulness, spatial localization, and compactness while also benefiting detection performance. From a unified perspective, the proposed framework is applicable to both single-scale and multi-scale DETR architectures.

Overall, the main contributions of this paper are as follows:
\begin{enumerate}
	\item We propose an end-to-end instance-specific visual explanation framework, termed EIVE, which models the cross-attention mechanism in DETR-like decoders as an instance-level feature attribution pathway, thereby deriving a corresponding saliency map for each detected instance from the forward-pass outputs.
	\item We introduce the CLHCF module, which selects a representative query for each target instance, tracks it across decoder layers, and performs adaptive hybrid fusion to construct compact and stable explanations.
	\item As a training-oriented application of the proposed framework, we propose AAJTS to regularize the spatial distribution of cross-attention, yielding more faithful and compact saliency maps while further improving detection performance.
	\item The proposed method exhibits strong architectural generality and can be consistently applied to both single-scale and multi-scale DETR-like object detectors. Its effectiveness is validated on multiple representative DETR-like frameworks.
	\item Experimental results on MS COCO 2017 (COCO), ExDark, and Cityscapes show that the proposed method improves the quality of instance-level saliency maps and achieves performance comparable to, or even better than, existing post-hoc visual explanation methods across a variety of mainstream evaluation metrics.
\end{enumerate}

\section{Related Work}
\subsection{Post-hoc Visual Explanation Methods}
Existing research on explainability has largely focused on the paradigm of post-hoc visual explanation, in which model predictions are analyzed after training by introducing an additional explainer, without modifying the original network architecture or training procedure. In image classification, this line of research has developed into a relatively mature framework, including class activation map-based methods~\cite{ref_CAM, ref_Grad-CAM, ref_LayerCAM}, input perturbation-based methods~\cite{ref_LIME, ref_RISE}, and SHAP-based methods~\cite{ref_SHAP, ref_DeepSHAP}.

In the field of object detection, researchers have gradually extended these post-hoc explanation methods to detector interpretation in order to accommodate the multi-instance prediction setting. For example, D-RISE~\cite{ref_D-RISE} generates saliency maps through random masking and extends RISE~\cite{ref_RISE} to object detection for revealing regions that are critical to detection outcomes. ODAM~\cite{ref_ODAM}, inspired by Grad-CAM~\cite{ref_Grad-CAM}, produces instance-specific explanations at the bounding-box level, and better captures two types of explanation requirements, namely object specification and object discrimination. In addition, D-HSIC~\cite{ref_D-HSIC} introduces a dependency-based measure grounded in global sensitivity analysis, improving the efficiency and effectiveness of black-box explanations. VPS~\cite{ref_VPS}, from the perspective of input-level feature confusion, proposes a black-box explanation method that improves detection performance using fewer critical regions while also revealing input-level factors that lead to detection failures. Overall, these post-hoc methods are highly flexible and can be adapted to different detectors. However, they still rely heavily on saliency visualization, are susceptible to noise, and remain limited in modeling feature confusion and higher-order interactions in the input.

In general, post-hoc visual explanation methods offer strong flexibility and can be applied to a wide range of detection models. Nevertheless, they usually require additional gradient computation or repeated input perturbations, resulting in substantial computational overhead. Moreover, since the explanation process is decoupled from the model’s internal reasoning path, the resulting explanations are often sensitive to noise and may lack stability and consistency.

\subsection{Interpretability of Transformer Models}
Transformer~\cite{ref_Transformer} was originally proposed for natural language processing, where its core self-attention mechanism explicitly models dependencies among input elements. With the successful application of Vision Transformer~\cite{ref_ViT, ref_DeiT, ref_Swin, ref_PVT, ref_CvT} to visual tasks, attention-based modeling has been widely introduced into visual recognition problems such as image classification and object detection. Given that attention weights encode relationships among features, a growing number of studies~\cite{ref_Attention-Rollout, ref_Attention-Visualization, Chefer_2021_ICCV, ref_AttCAT, ref_TokenTM, ref_FaithfulnessViT, ref_DETD} have begun to investigate the interpretability of Transformer models.

The interpretability of Transformers was first explored in image classification. Early studies typically visualized self-attention weights directly to illustrate the image regions attended to by the model during prediction. However, subsequent research has shown that attention visualization alone often fails to faithfully reflect the true basis of model decisions. Attention Rollout~\cite{ref_Attention-Rollout} reconstructs the contribution of input tokens by modeling information propagation across multiple attention layers, thereby alleviating, to some extent, the semantic ambiguity of high-level attention. Beyond Attention Visualization~\cite{ref_Attention-Visualization} further addresses this issue by integrating attention weights, gradient information, and relevance propagation, achieving substantially better explanation quality than direct attention visualization on classification tasks and revealing more discriminative feature attribution pathways inside Transformers. In addition, Generic Attention-model Explainability~\cite{Chefer_2021_ICCV} further extends Transformer interpretability beyond pure self-attention models to more general architectures, including bi-modal Transformers with co-attention and encoder-decoder Transformers. By modeling relevance propagation across different attention interactions, it shows that attention-based explanations can also be generalized to cross-modal and encoder-decoder settings.

Overall, these studies suggest that interpreting Transformers requires modeling cross-layer information propagation rather than relying on raw attention alone. Our work is inspired by this line of research, but addresses a different problem setting. Existing methods are mainly designed for classification-style predictions, requiring additional propagation or gradient computation after inference. In contrast, EIVE exploits the structural property of DETR-like detectors that decoder cross-attention naturally serves as an instance-level attribution pathway, and unifies query-instance prediction, forward saliency generation, cross-layer fusion, and attention-aware training within an end-to-end detection framework.

% \section{Hierarchical Decoupled Heads with Prototype Learning}
\section{Proposed Method}
\subsection{Overall Network Architecture}
\begin{figure*}[!t]
	\centering
	\includegraphics[width=\textwidth]{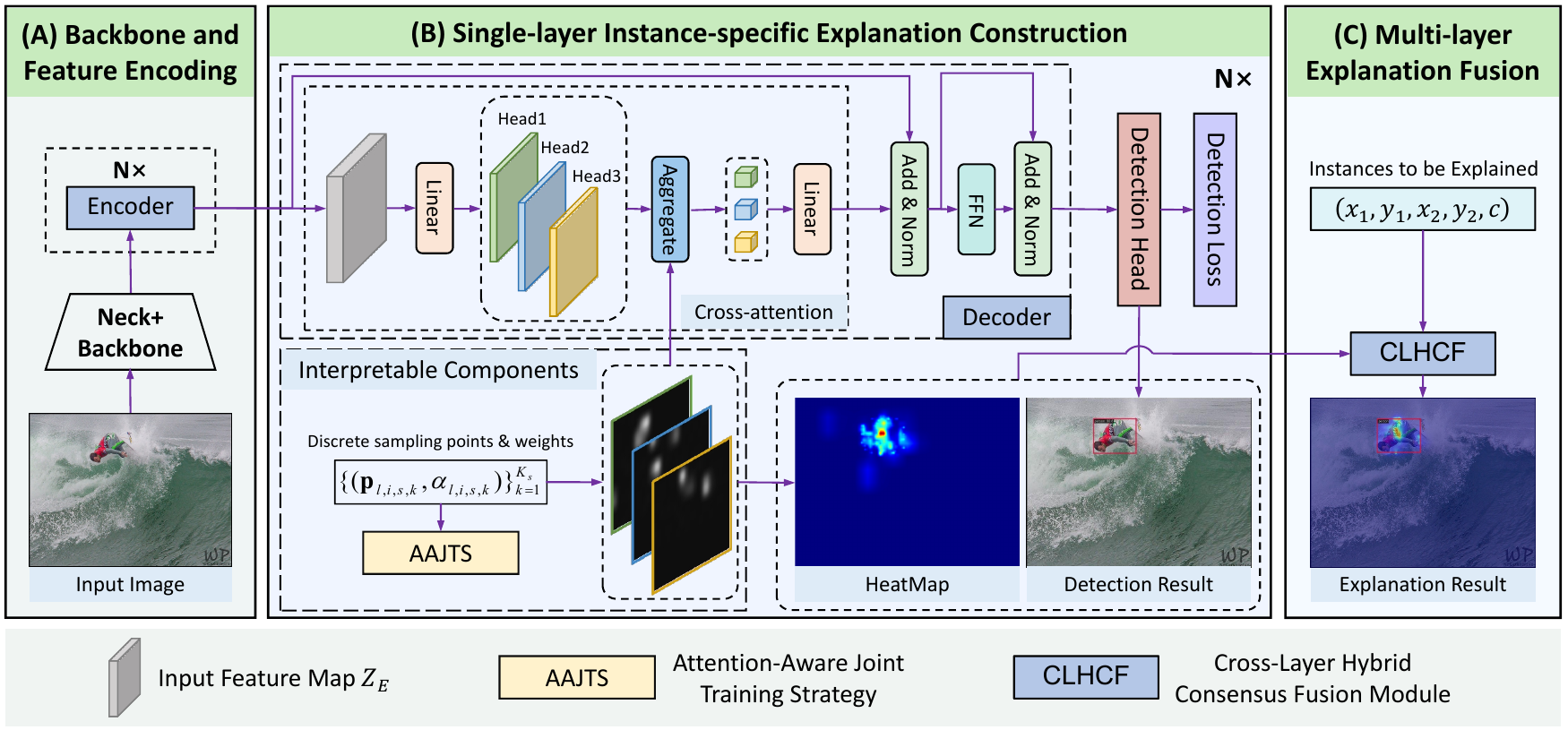}
	\caption{Overall architecture of the proposed EIVE.}
	\label{fig:Overall-architecture}
\end{figure*}

To address the issues of high explanation overhead and limited explanation consistency in existing explainability methods for object detection, we propose an end-to-end instance-specific visual explanation framework, termed EIVE. Built upon the standard Detection Transformer architecture, EIVE integrates instance-level explanation modeling, the CLHCF module, and the AAJTS into a unified framework, as illustrated in Fig.~\ref{fig:Overall-architecture}.

Given an input image, the model first extracts and encodes image features through the Backbone and Encoder. The Decoder then interacts with the encoded features through a set of object queries to produce object class and bounding box predictions. Different from conventional Detection Transformers that focus solely on detection outputs, we further explicitly model the cross-attention in the decoder as an instance-level explanation signal, which characterizes the relative contributions of different spatial regions to the prediction of a target instance. In this way, the model can derive the corresponding instance-level explanations directly from the forward-pass outputs after obtaining detection results, without requiring additional gradient computation or input perturbation.

\begin{figure*}[!t]
	\centering
	\includegraphics[width=\textwidth]{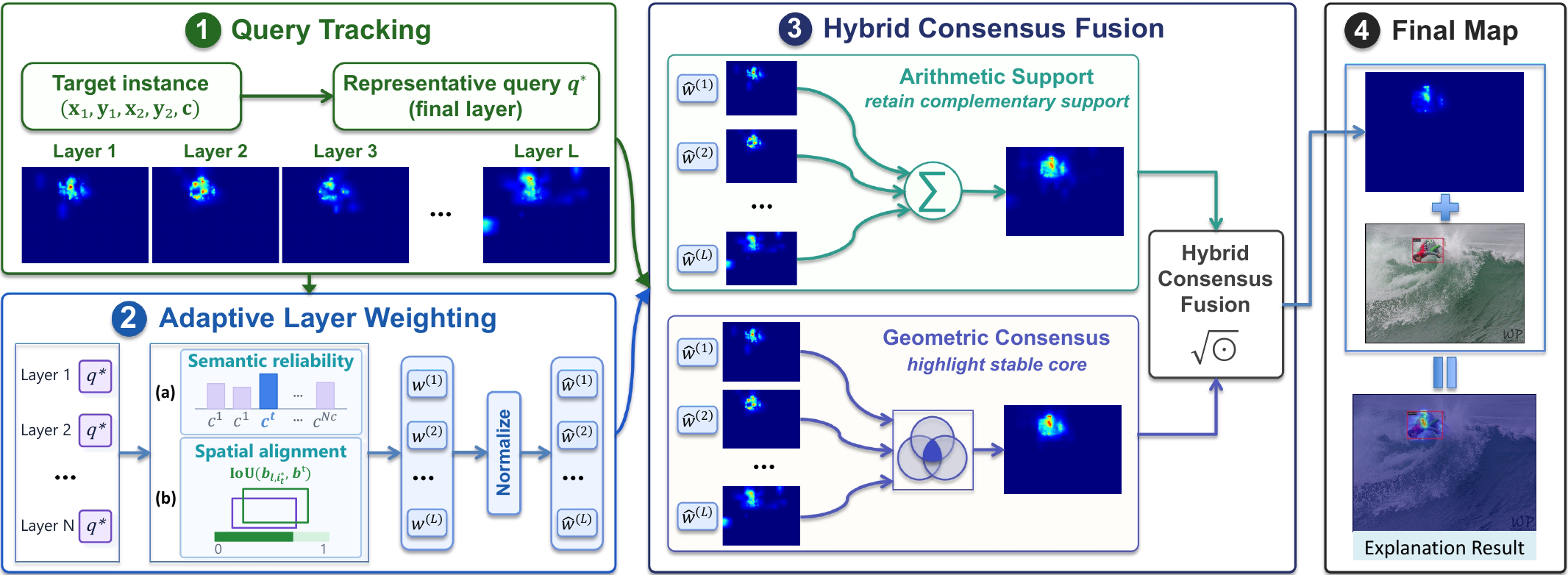}
	\caption{Cross-Layer Hybrid Consensus Fusion (CLHCF) Module.}
	\label{fig:CLHCF}
\end{figure*}

On this basis, we design the CLHCF module, whose structure is shown in Fig.~\ref{fig:CLHCF}. For a target instance, this module first selects a representative query from the last decoder layer and then tracks it across selected decoder layers. Based on the corresponding layer-wise saliency maps, CLHCF further performs adaptive cross-layer fusion with hybrid consensus modeling, thereby yielding explanations that are more compact, stable, and semantically consistent.

Furthermore, as a training-oriented application of the proposed explanation framework, we present the AAJTS. By imposing constraints on the spatial distribution of cross-attention, this strategy guides the model to learn more meaningful instance-level attribution representations, thereby improving the interpretability of the generated saliency maps while further benefiting detection performance.

\subsection{Instance-Level Modeling in Detection Transformers}
Detection Transformers formulate object detection as a set prediction problem. A fixed set of learnable object queries interacts with encoded image features and is directly transformed into instance-level predictions. This query-based decoding process provides a natural foundation for instance-specific explanation, because each object query is responsible for aggregating visual evidence for a potential object instance.

Given an input image $\mathbf{X} \in \mathbb{R}^{3 \times H \times W}$, the backbone extracts a feature map
$\mathbf{F} \in \mathbb{R}^{C \times H' \times W'}$, which is flattened into $N=H'W'$ visual tokens and further encoded by the Transformer encoder:
\begin{equation}
	\mathbf{Z}_E =
	\mathrm{Encoder}\bigl(
	\mathrm{Flatten}(\mathrm{Backbone}(\mathbf{X}))
	\bigr),
\end{equation}
where $\mathbf{Z}_E \in \mathbb{R}^{N \times C}$ denotes the encoded image representation.

The decoder maintains $M$ object queries:
\begin{equation}
	\mathbf{Q}_0=\{\mathbf{q}_1,\mathbf{q}_2,\ldots,\mathbf{q}_M\}, 
	\qquad \mathbf{q}_i \in \mathbb{R}^{C},
\end{equation}
where each query corresponds to a potential object instance. During decoding, object queries interact with encoded image features through cross-attention and progressively update their instance representations. Since the cross-attention weights explicitly describe how each query aggregates information from spatial image features, they can be naturally interpreted as instance-level spatial attribution signals. For clarity, we briefly introduce the cross-attention formulations in single-scale and multi-scale Detection Transformers.

\subsubsection{Cross-Attention Mechanism in Single-Scale Detection Transformers}
In single-scale Detection Transformers, each object query attends to all encoded spatial tokens. At the $l$-th decoder layer, the cross-attention operation can be written as:
\begin{equation}
	\mathbf{A}_l=
	\mathrm{softmax}\!\left(
	\frac{\mathbf{Q}_{l-1}\mathbf{K}_E^{\top}}{\sqrt{d}}
	\right), 
	\qquad
	\mathbf{H}_l=\mathbf{A}_l\mathbf{V}_E,
\end{equation}
where $\mathbf{K}_E=\mathbf{Z}_E\mathbf{W}_K$ and $\mathbf{V}_E=\mathbf{Z}_E\mathbf{W}_V$ are the projected key and value features, respectively. $\mathbf{A}_l \in \mathbb{R}^{M \times N}$ denotes the cross-attention matrix, $\mathbf{H}_l \in \mathbb{R}^{M \times C}$ denotes the updated query features, and ${d}$ is the head dimension.

For the $i$-th object query, the attention vector $\mathbf{A}_l^{(i)} \in \mathbb{R}^{N}$ indicates the contribution of all spatial tokens to this query. Since the encoded tokens correspond to spatial positions on the feature map, $\mathbf{A}_l^{(i)}$ can be reshaped into a two-dimensional response map, which provides an instance-specific spatial explanation for the prediction associated with query $i$.

\subsubsection{Cross-Attention Mechanism in Multi-Scale Detection Transformers}
Multi-scale Detection Transformers improve efficiency and multi-scale modeling ability by replacing global cross-attention with deformable cross-attention. Given multi-scale encoded features
$\{\mathbf{Z}_E^{(1)}, \mathbf{Z}_E^{(2)}, \ldots, \mathbf{Z}_E^{(S)}\}$,
where $\mathbf{Z}_E^{(s)} \in \mathbb{R}^{N_s \times C}$ denotes the encoded feature at scale $s$, each query only attends to a sparse set of sampling points across different scales.

For the $i$-th query at the $l$-th decoder layer, the cross-attention output is formulated as:
\begin{equation}
	\mathbf{h}_{i,l}
	=
	\sum_{s=1}^{S}\sum_{k=1}^{K}
	\alpha_{i,l,s,k}\cdot
	\mathbf{Z}_E^{(s)}\!\left(\mathbf{p}_{i,l,s,k}\right),
\end{equation}
where $\mathbf{p}_{i,l,s,k}$ denotes the $k$-th sampling location on the feature map of scale $s$, and $\alpha_{i,l,s,k}$ is the corresponding attention weight. The decoder output at layer $l$ is then:
\begin{equation}
	\mathbf{H}_l=
	[\mathbf{h}_{1,l}, \mathbf{h}_{2,l}, \ldots, \mathbf{h}_{M,l}]^{\top}
	\in \mathbb{R}^{M \times C}.
\end{equation}

Compared with single-scale global attention, deformable cross-attention defines a sparse but spatially explicit attention distribution over multi-scale feature maps. Therefore, the sampling locations and their associated attention weights can also be rearranged into scale-dependent response maps, which serve as the basis for constructing instance-level explanations. When $S=1$ and $K=H' \times W'$, this mechanism degenerates to single-scale cross-attention in form.

\subsection{Modeling Cross-Attention as Instance-Level Explanation Signals}
The cross-attention mechanism in Detection Transformers essentially characterizes how each object query aggregates information in the image space. For a given target instance, the prediction is not determined by a single spatial location. Instead, it is jointly contributed by a set of encoder features with explicit spatial structure under the modulation of cross-attention weights. Therefore, if the spatial distribution pattern of cross-attention can be explicitly modeled, it can be naturally interpreted as an instance-level explanation signal that reflects the relative importance of different spatial regions to the prediction of the target instance.

\subsubsection{Cross-Attention as Instance-Level Spatial Attribution}
\label{sec_CAILSA}
From a unified modeling perspective, we do not distinguish between single-scale and multi-scale Detection Transformers in terms of their implementation details. Instead, we assume that, at the $l$-th decoder layer, for the $i$-th object query, the cross-attention mechanism has determined a set of discrete spatial sampling locations and their corresponding attention weights. This assumption holds for both multi-scale and single-scale models: in the multi-scale case, the discrete sampling locations are implicitly determined by reference points and the interpolation mechanism. In the single-scale case, the sampling locations naturally cover all discrete spatial positions on the entire feature map.

Specifically, at the $l$-th decoder layer, for the $i$-th object query $\mathbf{q}_{i,l}$, the cross-attention mechanism produces, on the feature map of the $s$-th scale, a set of discrete sampling points and their associated weights:
\begin{equation}
	\left\{\left(\mathbf{p}_{i,l,s,k}, \alpha_{i,l,s,k}\right)\right\}_{k=1}^{K_s},
	\label{eq:cross_attn_sampling}
\end{equation}
where $\mathbf{p}_{i,l,s,k} = \left(x_{i,l,s,k}, y_{i,l,s,k}\right)$ denotes the $k$-th pixel location on the feature map at scale $s$, $\alpha_{i,l,s,k}$ represents the contribution strength of this location to the feature aggregation of instance $\mathbf{q}_{i,l}$, and $K_s$ denotes the number of sampling points, which is typically a small constant in multi-scale models, while in single-scale global attention it equals the total number of spatial locations on the feature map.

Based on the above representation, an instance-level explanatory map $\mathbf{E}_{i,l}^{(s)} \in \mathbb{R}^{H_s \times W_s}$ can be constructed for $\mathbf{q}_{i,l}$ at scale $s$. This explanatory map is initialized as an all-zero tensor and is obtained by accumulating all sampling points:
\begin{equation}
	\mathbf{E}_{i,l}^{(s)}(y,x)
	=
	\sum_{k=1}^{K_s}
	\alpha_{i,l,s,k}\cdot
	\mathbb{I}\bigl((y,x)=\mathbf{p}_{i,l,s,k}\bigr),
\end{equation}
where $\mathbb{I}(\cdot)$ is the indicator function used to assign attention weights to the corresponding discrete spatial locations. This process is essentially equivalent to explicitly rearranging the spatial allocation of cross-attention into a two-dimensional response map, thereby characterizing the contribution of different spatial regions to the current target instance.

Since the spatial sizes of different scales are inconsistent, they cannot be fused directly. We therefore first perform spatial alignment on the explanatory maps at different scales. Specifically, we introduce a linear interpolation operator $\mathcal{I}_{\text{bilinear}}(\cdot)$ to map the scale-specific explanatory maps to the same spatial resolution $H \times W$ as the input image, and then perform cross-scale fusion on this basis. The multi-scale fused instance-level explanatory map at the $l$-th decoder layer is uniformly defined as:
\begin{equation}
	\mathbf{E}_{i,l}
	=
	\frac{1}{S}\sum_{s=1}^{S}
	\mathcal{I}_{\text{bilinear}}\bigl(\mathbf{E}_{i,l}^{(s)}; (H,W)\bigr),
\end{equation}
where $S$ denotes the total number of feature scales. This fusion process introduces no additional parameters and only reflects the overall attention distribution of cross-attention across different scales.

\subsubsection{Cross-Layer Hybrid Consensus Fusion (CLHCF) Module} 
Based on the instance-level spatial attribution maps $\mathbf{E}_{i,l}$ defined in Sec.~\ref{sec_CAILSA}, we further aggregate explanation signals across decoder layers to obtain a stable explanation map for a target instance. Direct layer-wise fusion is suboptimal because different decoder layers may focus on different spatial patterns of the same instance, such as object core, local parts, or surrounding context. To reduce such cross-layer drift, we first select a single representative query from the last decoder layer and then track it across layers, so that the fusion remains centered on the same target instance.

To provide a unified criterion for both query selection and layer weighting, we define a query--instance relevance function:
\begin{equation}
	\phi(i,l,t)
	=
	o_{l,i}(c^{t})
	\cdot
	\mathrm{IoU}(\mathbf{b}_{l,i}, \mathbf{b}^{t}),
\end{equation}
where $c^{t}$ and $\mathbf{b}^{t}$ denote the target category and target box of the instance $t$ to be explained, respectively. They specify the object-level decision of interest and can be provided by the user or determined by the evaluation protocol. $o_{l,i}(c^{t})$ is the predicted probability of class $c^{t}$ for the $i$-th query at decoder layer $l$, and $\mathbf{b}_{l,i}$ is the corresponding predicted box. This relevance function jointly measures the semantic consistency and localization consistency between a query and the target instance.

Based on this relevance function, the representative query is selected from the last decoder layer by:
\begin{equation}
	i_t^{*} = \arg\max_{i} \phi(i,L,t),
\end{equation}
where $i_t^{*}$ denotes the index of the most relevant query for instance $t$ at the final decoding stage. The selected query is then reused in all layers involved in fusion.

For each selected decoder layer, we use the attribution map of the same query $i_t^{*}$ and normalize it into a non-negative spatial distribution, denoted by $\mathbf{M}_{l}^{t}$, so that layer-wise responses become comparable and the fusion focuses on spatial patterns rather than raw response magnitudes.

Since different decoder layers are not equally reliable, we further introduce an adaptive layer weighting scheme:
\begin{equation}
	w_{l}^{t}
	=
	\phi(i_t^{*},l,t),
\end{equation}
where $w_{l}^{t}$ is the layer reliability weight. The weights are clamped to be non-negative and normalized across selected layers to obtain $\bar{w}_{l}^{t}$.

To obtain the final explanation, we adopt a hybrid consensus fusion strategy with two complementary branches. The arithmetic branch preserves complementary target-related responses across layers:
\begin{equation}
	\mathbf{S}_{\mathrm{arith}}^{t}
	=
	\sum_{l=L-L'+1}^{L} \bar{w}_{l}^{t}\,\mathbf{M}_{l}^{t},
\end{equation}
where $\mathbf{S}_{\mathrm{arith}}^{t}$ denotes the arithmetic fusion result and $L'$ denotes the number of participating layers. This branch behaves as a weighted mixture of layer-wise attribution maps and is effective for retaining target-related regions that may appear differently across decoder layers.

The geometric branch emphasizes regions that are consistently highlighted across layers:
\begin{equation}
	\mathbf{S}_{\mathrm{geom}}^{t}
	=
	\exp\left(
	\sum_{l=L-L'+1}^{L} \bar{w}_{l}^{t}
	\log(\mathbf{M}_{l}^{t}+\epsilon)
	\right),
\end{equation}
where $\mathbf{S}_{\mathrm{geom}}^{t}$ denotes the geometric fusion result. Different from arithmetic fusion, the geometric branch assigns low responses to regions that are activated only in a few layers, thereby suppressing unstable background responses and scattered contextual activations. Since target-related regions are more likely to be repeatedly attended by the tracked query across decoder layers, this consensus constraint encourages the fused saliency map to concentrate on the target instance, leading to improved instance-level focus and better spatial localization ability.

Finally, the two branches are combined as:
\begin{equation}
	\mathbf{S}^{t}
	=
	\sqrt{
		\mathbf{S}_{\mathrm{arith}}^{t}
		\odot
		\mathbf{S}_{\mathrm{geom}}^{t}
		+\epsilon
	},
\end{equation}
where $\odot$ denotes element-wise multiplication and $\mathbf{S}^{t}$ is the hybrid fusion result. The square-root operation stabilizes the dynamic range of the fused map. We then apply min-max normalization to $\mathbf{S}^{t}$ to obtain the final explanation map for visualization and evaluation.

\subsection{Attention-Aware Joint Training Strategy}
As a training-oriented application of the proposed framework, we present an Attention-Aware Joint Training Strategy (AAJTS). In Detection Transformers, each object query relies on cross-attention to aggregate visual features for classification and localization. However, during standard detector training, the spatial allocation of cross-attention is only indirectly optimized by the final detection loss. As a result, a matched query may still attend to irrelevant background regions or weakly related contextual responses, which can introduce noisy features into the instance representation and weaken target-specific feature aggregation. To mitigate this issue, AAJTS uses bounding boxes as weak spatial priors to regularize the attention mass distribution of matched queries, rather than treating them as precise attention masks. This strategy encourages the model to allocate more attention mass to target-related regions while softly suppressing far-background responses. As a result, the model learns more stable and concentrated attribution representations, leading to more faithful and compact explanations while further benefiting detection performance.

Following the Hungarian matching assignment in the detection branch, let $m_l(t)$ denote the index of the query matched to the $t$-th ground-truth instance at the $l$-th decoder layer, and let $\mathbf{b}_{gt}$ be its bounding box. That is, the corresponding matched query is $\mathbf{q}_{m_l(t),l}$. Based on the cross-attention sampling representation defined in Eq.~\eqref{eq:cross_attn_sampling}, the sampling location and attention weight of this matched query at scale $s$ are denoted as $\mathbf{p}_{m_l(t),l,s,k}$ and $\alpha_{m_l(t),l,s,k}$, respectively. For attention mass computation, we normalize the attention weights of each matched query across all scales and sampling points, and denote the normalized weight as $\tilde{\alpha}_{l,t,s,k}$. When multi-head attention is used, the weights are first averaged over heads before normalization.

For a spatial location $\mathbf{p}=(p_x,p_y)$ and a box $\mathbf{b}$ with center $(c_x,c_y)$, width $w$, and height $h$, we define the normalized distance from $\mathbf{p}$ to $\mathbf{b}$ as:
\begin{equation}
	d(\mathbf{p},\mathbf{b})
	=
	\left\|
	\left(
	\frac{\left[|p_x-c_x|-\frac{w}{2}\right]_+}{\frac{w}{2}+\epsilon},
	\frac{\left[|p_y-c_y|-\frac{h}{2}\right]_+}{\frac{h}{2}+\epsilon}
	\right)
	\right\|_2 ,
\end{equation}
where $[x]_+=\max(x,0)$. The distance is zero inside the box and increases smoothly outside it.

Let $\mathbf{b}_{gt}^s$ denote the projection of the ground-truth box $\mathbf{b}_{gt}$ onto the feature map at scale $s$. We then define the target-related soft prior and the far-background prior as:
\begin{equation}
	\psi_t^s(\mathbf{p})
	=
	\exp
	\left(
	-
	d(\mathbf{p},\mathbf{b}_{gt}^s)^2
	\right),
	\,
	\rho_t^s(\mathbf{p})
	=
	\left(
	1-\psi_t^s(\mathbf{p})
	\right)^2 ,
\end{equation}
where $\psi_t^s(\mathbf{p})$ assigns higher scores to locations inside or near the target box, while $\rho_t^s(\mathbf{p})$ gives larger penalties to far-background regions. Using the above priors, the target-related attention mass and far-background attention mass of the matched query are defined as:
\begin{equation}
	\begin{aligned}
		M_{\mathrm{rel}}^{(l,t)}
		&=
		\sum_{s=1}^{S}
		\sum_{k=1}^{K_s}
		\tilde{\alpha}_{l,t,s,k}
		\psi_t^s
		\left(
		\mathbf{p}_{m_l(t),l,s,k}
		\right), \\
		M_{\mathrm{far}}^{(l,t)}
		&=
		\sum_{s=1}^{S}
		\sum_{k=1}^{K_s}
		\tilde{\alpha}_{l,t,s,k}
		\rho_t^s
		\left(
		\mathbf{p}_{m_l(t),l,s,k}
		\right).
	\end{aligned}
\end{equation}

Instead of imposing point-wise hard supervision on sampling locations, AAJTS constrains the relative distribution of aggregate attention mass by encouraging a larger fraction of attention to be assigned to target-related regions. The final AAJTS loss is averaged over all decoder layers and matched ground-truth instances:
\begin{equation}
	\mathcal{L}_{\mathrm{AAJTS}}
	=
	\frac{1}{L}
	\sum_{l=1}^{L}
	\frac{1}{|\mathcal{T}_l|}
	\sum_{t\in\mathcal{T}_l}
	-\log
	\left(
	\frac{
		M_{\mathrm{rel}}^{(l,t)}+\epsilon
	}{
		M_{\mathrm{rel}}^{(l,t)}
		+
		M_{\mathrm{far}}^{(l,t)}
		+\epsilon
	}
	\right),
\end{equation}
where \(\mathcal{T}_l\) denotes the set of ground-truth instances matched at the \(l\)-th decoder layer. This objective only regularizes the relative attention mass between target-related and far-background regions, while the learning of discriminative object features remains driven by the original detection loss. The overall training objective is:
\begin{equation}
	\mathcal{L}
	=
	\mathcal{L}_{\mathrm{det}}
	+
	\mathcal{L}_{\mathrm{AAJTS}},
\end{equation}
where $\mathcal{L}_{\mathrm{det}}$ denotes the original loss of the detector.

\begin{table}[!t]
	\caption{Detection performance of the pretrained object detection models on the COCO, ExDark, and Cityscapes datasets. Input image sizes are shown below the dataset names.}
	\label{tab:detector_performance}
	\centering
	\renewcommand{\arraystretch}{1.1}
	\resizebox{\columnwidth}{!}{
		\begin{tabular}{@{\hspace{0.6em}}
				>{\arraybackslash}m{1.2cm}
				>{\arraybackslash}m{2.2cm} |
				>{\centering\arraybackslash}m{0.5cm}
				>{\centering\arraybackslash}m{0.75cm}
				>{\centering\arraybackslash}m{0.75cm}
				>{\centering\arraybackslash}m{0.7cm}
				>{\centering\arraybackslash}m{0.7cm}
				>{\centering\arraybackslash}m{0.7cm}
				@{\hspace{0.6em}}}
			\hline
			\textbf{Datasets} & \textbf{Detectors} & $\mathbf{mAP}$ & $\mathbf{mAP}_{50}$ & $\mathbf{mAP}_{75}$ & $\mathbf{mAP}_{s}$ & $\mathbf{mAP}_{m}$ & $\mathbf{mAP}_{l}$ \\
			\hline
			\multirow[c]{5}{*}{\shortstack[c]{COCO\\ \footnotesize(1333$\times$800)}}
			& DETR              & 0.399 & 0.604 & 0.417 & 0.176 & 0.435 & 0.594 \\
			& Conditional-DETR  & 0.410 & 0.619 & 0.435 & 0.204 & 0.445 & 0.599 \\
			& DAB-DETR          & 0.423 & 0.629 & 0.452 & 0.216 & 0.461 & 0.613 \\
			& Deformable-DETR   & 0.470 & 0.661 & 0.509 & 0.302 & 0.497 & 0.620 \\
			& DINO              & 0.490 & 0.664 & 0.533 & 0.314 & 0.522 & 0.640 \\
			\hline
			\multirow[c]{5}{*}{\shortstack[c]{ExDark\\ \footnotesize(1333$\times$800)}}
			& DETR              & 0.345 & 0.635 & 0.325 & 0.051 & 0.212 & 0.427 \\
			& Conditional-DETR  & 0.362 & 0.677 & 0.348 & 0.078 & 0.264 & 0.434 \\
			& DAB-DETR          & 0.368 & 0.669 & 0.368 & 0.105 & 0.266 & 0.439 \\
			& Deformable-DETR   & 0.460 & 0.748 & 0.502 & 0.162 & 0.366 & 0.520 \\
			& DINO              & 0.476 & 0.745 & 0.527 & 0.125 & 0.366 & 0.544 \\
			\hline
			\multirow[c]{5}{*}{\shortstack[c]{Cityscapes\\ \footnotesize(2048$\times$1024)}}
			& DETR              & 0.312 & 0.566 & 0.293 & 0.071 & 0.300 & 0.554 \\
			& Conditional-DETR  & 0.337 & 0.591 & 0.324 & 0.086 & 0.327 & 0.563 \\
			& DAB-DETR          & 0.363 & 0.618 & 0.355 & 0.104 & 0.370 & 0.593 \\
			& Deformable-DETR   & 0.389 & 0.656 & 0.392 & 0.159 & 0.388 & 0.587 \\
			& DINO              & 0.430 & 0.654 & 0.435 & 0.191 & 0.441 & 0.655 \\
			\hline
		\end{tabular}
	}
\end{table}

\begin{table*}[!t]
	\caption{Quantitative comparison results of different visual explanation methods on the COCO dataset. Time (s) denotes the average explanation time required for a single target instance. \textcolor{red}{Red} and \textcolor{blue}{blue} indicate the best and second-best results, respectively.}
	\label{tab:coco_explanation_comparison}
	\centering
	\renewcommand{\arraystretch}{1.1}
	\resizebox{\textwidth}{!}{
		\begin{tabular}{@{\hspace{0.6em}}
				>{\arraybackslash}m{2.0cm}
				>{\arraybackslash}m{1.5cm} |
				>{\centering\arraybackslash}m{1.5cm}
				>{\centering\arraybackslash}m{1.5cm}
				>{\centering\arraybackslash}m{1.3cm}
				>{\centering\arraybackslash}m{1.3cm} |
				>{\centering\arraybackslash}m{1.7cm}
				>{\centering\arraybackslash}m{1.6cm}
				>{\centering\arraybackslash}m{1.3cm} |
				>{\centering\arraybackslash}m{1.3cm}
				@{\hspace{0.6em}}}
			\hline
			\textbf{Detectors} & \textbf{Methods} &
			\textbf{Ins.\ Class $\uparrow$} &
			\textbf{Del.\ Class $\downarrow$} &
			\textbf{Ins.\ IoU $\uparrow$} &
			\textbf{Del.\ IoU $\downarrow$} &
			\textbf{Point Game $\uparrow$} &
			\textbf{Energy PG $\uparrow$} &
			\textbf{Comp.\ $\downarrow$} &
			\textbf{Time (s) $\downarrow$} \\
			\hline
			\multirow{6}{*}{{DETR}}
			& GradCAM & 0.0975 & 0.9660 & 0.1042 & 0.9587 & 0.1346 & 0.1082 & 7.8037 & \textcolor{blue}{0.1662} \\
			& GradCAM++ & 0.1486 & 0.9158 & 0.1529 & 0.9246 & 0.1923 & 0.1343 & 7.0832 & 0.1734 \\
			& ODAM & 0.1777 & \textcolor{blue}{0.7677} & 0.1961 & 0.7495 & \textcolor{blue}{0.5976} & \textcolor{blue}{0.2567} & \textcolor{blue}{5.8981} & 0.1747 \\
			& D-RISE & 0.1771 & 0.7731 & 0.1962 & \textcolor{blue}{0.7336} & 0.5931 & 0.0944 & 7.7725 & 17.9977 \\
			& VPS & \textcolor{red}{0.2382} & 0.7959 & \textcolor{blue}{0.2307} & 0.7349 & 0.2769 & 0.1726 & 7.6499 & 62.9709 \\
			& EIVE (ours) & \textcolor{blue}{0.2287} & \textcolor{red}{0.7295} & \textcolor{red}{0.2444} & \textcolor{red}{0.6881} & \textcolor{red}{0.8475} & \textcolor{red}{0.7736} & \textcolor{red}{0.8389} & \textcolor{red}{0.0440} \\
			\hline
			\multirow{6}{*}{{Conditional-DETR}}
			& GradCAM & 0.0736 & 0.3445 & 0.0393 & 0.1170 & 0.1867 & 0.1331 & 6.0099 & \textcolor{blue}{0.1685} \\
			& GradCAM++ & 0.1093 & 0.3187 & 0.0452 & 0.1229 & 0.1911 & 0.1534 & 6.0054 & 0.1708 \\
			& ODAM & 0.1704 & \textcolor{red}{0.2900} & 0.0727 & 0.1163 & \textcolor{blue}{0.7742} & \textcolor{blue}{0.3498} & \textcolor{blue}{4.2487} & 0.1732 \\
			& D-RISE & 0.1571 & \textcolor{blue}{0.2938} & 0.0687 & 0.1162 & 0.7353 & 0.1160 & 5.9600 & 30.0825 \\
			& VPS & \textcolor{blue}{0.2057} & 0.2977 & \textcolor{blue}{0.0778} & \textcolor{red}{0.1120} & 0.4040 & 0.2267 & 5.6309 & 66.1161 \\
			& EIVE (ours) & \textcolor{red}{0.2235} & 0.2970 & \textcolor{red}{0.0881} & \textcolor{blue}{0.1152} & \textcolor{red}{0.8872} & \textcolor{red}{0.7952} & \textcolor{red}{0.8778} & \textcolor{red}{0.0512} \\
			\hline
			\multirow{6}{*}{{DAB-DETR}}
			& GradCAM & 0.0269 & 0.1956 & 0.0068 & 0.0845 & 0.1566 & 0.1446 & 6.2886 & \textcolor{blue}{0.1629} \\
			& GradCAM++ & 0.0329 & 0.1879 & 0.0074 & 0.0861 & 0.1368 & 0.1477 & 6.6473 & 0.1687 \\
			& ODAM & 0.0449 & 0.1743 & 0.0210 & 0.0814 & 0.6829 & \textcolor{blue}{0.3210} & \textcolor{blue}{4.5818} & 0.1735 \\
			& D-RISE & 0.0460 & \textcolor{red}{0.1728} & \textcolor{blue}{0.0237} & \textcolor{blue}{0.0805} & \textcolor{blue}{0.7169} & 0.1136 & 6.3754 & 29.0599 \\
			& VPS & \textcolor{red}{0.0608} & \textcolor{blue}{0.1742} & 0.0193 & \textcolor{red}{0.0800} & 0.3452 & 0.2258 & 5.8409 & 69.2042 \\
			& EIVE (ours) & \textcolor{blue}{0.0550} & 0.1799 & \textcolor{red}{0.0271} & \textcolor{blue}{0.0805} & \textcolor{red}{0.9047} & \textcolor{red}{0.8120} & \textcolor{red}{1.2089} & \textcolor{red}{0.0499} \\
			\hline
			\multirow{6}{*}{{Deformable-DETR}}
			& GradCAM & 0.0869 & 0.4639 & 0.0511 & 0.2813 & 0.2014 & 0.1326 & 6.5896 & \textcolor{blue}{0.1999} \\
			& GradCAM++ & 0.1196 & 0.4300 & 0.0699 & 0.2759 & 0.2300 & 0.1392 & 6.7856 & 0.2261 \\
			& ODAM & 0.2725 & 0.3321 & \textcolor{blue}{0.2865} & 0.1810 & \textcolor{blue}{0.7506} & \textcolor{blue}{0.3010} & \textcolor{blue}{4.7976} & 0.2296 \\
			& D-RISE & 0.2699 & \textcolor{blue}{0.3195} & 0.2749 & 0.1639 & 0.6896 & 0.1010 & 7.3167 & 42.4213 \\
			& VPS & \textcolor{blue}{0.3102} & \textcolor{red}{0.3139} & 0.2549 & \textcolor{red}{0.1530} & 0.3324 & 0.2133 & 6.9094 & 110.9404 \\
			& EIVE (ours) & \textcolor{red}{0.3872} & 0.3208 & \textcolor{red}{0.4765} & \textcolor{blue}{0.1615} & \textcolor{red}{0.9956} & \textcolor{red}{0.9163} & \textcolor{red}{0.5115} & \textcolor{red}{0.0763} \\
			\hline
			\multirow{6}{*}{{DINO}}
			& GradCAM & 0.0754 & 0.4949 & 0.0632 & 0.3114 & 0.2099 & 0.1376 & 6.0639 & \textcolor{blue}{0.2214} \\
			& GradCAM++ & 0.1219 & 0.4465 & 0.0895 & 0.3027 & 0.2530 & 0.1541 & 6.1136 & 0.2446 \\
			& ODAM & 0.2780 & 0.3248 & \textcolor{blue}{0.3777} & 0.1788 & \textcolor{red}{0.8192} & \textcolor{blue}{0.3468} & \textcolor{blue}{4.2415} & 0.2542 \\
			& D-RISE & 0.2464 & \textcolor{blue}{0.3192} & 0.3269 & 0.1657 & 0.7202 & 0.1110 & 6.6050 & 55.6982 \\
			& VPS & \textcolor{blue}{0.2808} & \textcolor{red}{0.3167} & 0.2872 & \textcolor{blue}{0.1591} & 0.3498 & 0.2509 & 5.9937 & 118.3257 \\
			& EIVE (ours) & \textcolor{red}{0.3179} & 0.3221 & \textcolor{red}{0.4459} & \textcolor{red}{0.1559} & \textcolor{blue}{0.7263} & \textcolor{red}{0.7591} & \textcolor{red}{0.9727} & \textcolor{red}{0.0776} \\
			\hline
		\end{tabular}
	}
\end{table*}
\begin{figure*}[!t]
	\centering
	\includegraphics[width=\textwidth]{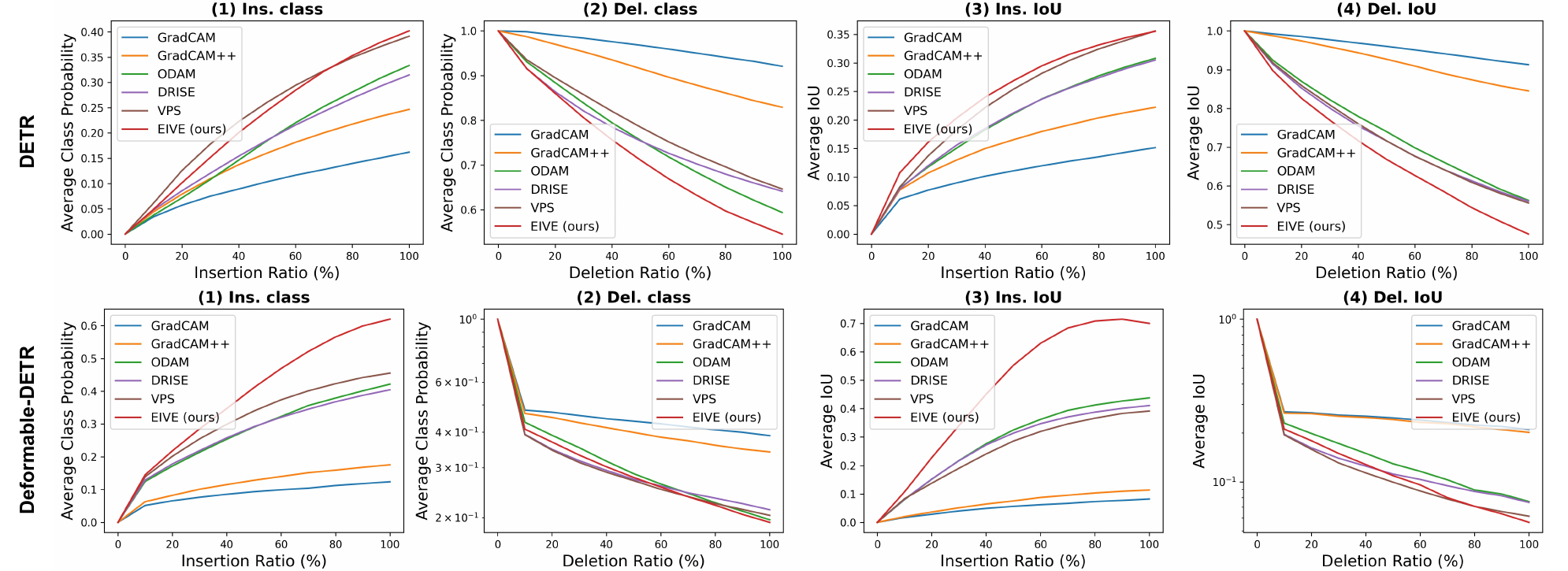}
	\caption{Average Insertion and Deletion Curves for Faithfulness Evaluation on the COCO Dataset.}
	\label{fig:Insert-delete-curves-coco}
\end{figure*}

\section{Experiments}
\subsection{Experimental Setup}
\subsubsection{Datasets}
We evaluate the proposed method on three representative object detection datasets, namely COCO~\cite{ref_COCO}, ExDark~\cite{ref_Exdark}, and Cityscapes~\cite{ref_Cityscapes}.

COCO is a large-scale general-purpose object detection dataset covering 80 common object categories, with rich scene variations and diverse object scale distributions. It contains 118,287 training images and 5,000 validation images.

ExDark is an object detection dataset designed for low-light environments. It contains 7,363 real-world images covering 12 object categories. Following the common experimental setting, we randomly split the dataset into 5,896 training images and 1,467 validation images.

Cityscapes is a large-scale urban scene dataset collected from real-world street environments, featuring densely distributed traffic participants and roadside objects. In our experiments, we use 8 object categories. The dataset contains 2,975 training images and 500 validation images.

\subsubsection{Models for Explanation}
To verify the applicability of the proposed method across different Detection Transformer architectures, we select five Transformer-based object detectors as the subject models. These include three single-scale detectors, namely DETR~\cite{ref_DETR}, Conditional DETR~\cite{ref_conditional}, and DAB-DETR~\cite{ref_dab}, as well as two multi-scale detectors, i.e., Deformable DETR~\cite{ref_deformable} and DINO~\cite{ref_dino}.

\subsubsection{Baselines}
We compare the proposed method with several representative explainability methods spanning different explanation paradigms. Specifically, the baselines include three gradient-based explanation methods, namely Grad-CAM~\cite{ref_Grad-CAM}, Grad-CAM++~\cite{ref_Grad-CAM++}, and ODAM~\cite{ref_ODAM}, as well as two perturbation-based methods, i.e., D-RISE~\cite{ref_D-RISE} and VPS~\cite{ref_VPS}. For all gradient-based methods, including Grad-CAM, Grad-CAM++, and ODAM, we use the same target layer for fair comparison, namely the output of the last backbone stage. For perturbation-based methods, D-RISE uses 2000 random masks, while VPS divides the image into 50 sub-regions for explanation generation.

\subsubsection{Implementation Details}
All experiments are conducted on a server with 8 NVIDIA GeForce RTX 3090 GPUs, using Python 3.9 and PyTorch 2.1.0. To ensure reproducibility, the random seed is set to 5. In the CLHCF module, the number of decoder layers participating in the final saliency fusion is set to $L'=6$, which is equal to the total number of decoder layers used in the detector. To interpret model decisions, we set the ground-truth bounding box and category of the object to be explained as the target box $\mathbf{b}^{t}$ and target category $c^{t}$ for all explanation methods.

\begin{table*}[!t]
	\caption{Quantitative comparison results of different visual explanation methods on the ExDark dataset. Time (s) denotes the average explanation time required for a single target instance. \textcolor{red}{Red} and \textcolor{blue}{blue} indicate the best and second-best results, respectively.}
	\label{tab:exdark_explanation_comparison}
	\centering
	\renewcommand{\arraystretch}{1.1}
	\resizebox{\textwidth}{!}{
		\begin{tabular}{@{\hspace{0.6em}}
				>{\arraybackslash}m{2.0cm}
				>{\arraybackslash}m{1.5cm} |
				>{\centering\arraybackslash}m{1.5cm}
				>{\centering\arraybackslash}m{1.5cm}
				>{\centering\arraybackslash}m{1.3cm}
				>{\centering\arraybackslash}m{1.3cm} |
				>{\centering\arraybackslash}m{1.7cm}
				>{\centering\arraybackslash}m{1.6cm}
				>{\centering\arraybackslash}m{1.3cm} |
				>{\centering\arraybackslash}m{1.3cm}
				@{\hspace{0.6em}}}
			\hline
			\textbf{Detectors} & \textbf{Methods} &
			\textbf{Ins.\ Class $\uparrow$} &
			\textbf{Del.\ Class $\downarrow$} &
			\textbf{Ins.\ IoU $\uparrow$} &
			\textbf{Del.\ IoU $\downarrow$} &
			\textbf{Point Game $\uparrow$} &
			\textbf{Energy PG $\uparrow$} &
			\textbf{Comp.\ $\downarrow$} &
			\textbf{Time (s) $\downarrow$} \\
			\hline
			\multirow{6}{*}{{DETR}}
			& GradCAM & 0.1103 & 0.9031 & 0.1488 & 0.9276 & 0.2043 & 0.1317 & 4.3269 & \textcolor{blue}{0.1609} \\
			& GradCAM++ & 0.2414 & 0.7422 & 0.2660 & 0.8367 & 0.3263 & 0.1786 & 3.8839 & 0.1733 \\
			& ODAM & 0.3008 & 0.6296 & 0.3088 & 0.7008 & 0.7339 & \textcolor{blue}{0.3285} & \textcolor{blue}{3.2941} & 0.1786 \\
			& D-RISE & \textcolor{blue}{0.3669} & \textcolor{blue}{0.6053} & \textcolor{blue}{0.3488} & 0.6708 & \textcolor{blue}{0.8399} & 0.0995 & 4.3755 & 21.1535 \\
			& VPS & 0.3426 & 0.6171 & 0.3286 & \textcolor{blue}{0.6571} & 0.4001 & 0.1761 & 4.2663 & 61.0919 \\
			& EIVE (ours) & \textcolor{red}{0.3924} & \textcolor{red}{0.6029} & \textcolor{red}{0.3815} & \textcolor{red}{0.6568} & \textcolor{red}{0.9437} & \textcolor{red}{0.9008} & \textcolor{red}{0.7049} & \textcolor{red}{0.0472} \\
			\hline
			\multirow{6}{*}{{Deformable-DETR}}
			& GradCAM & 0.0578 & 0.5463 & 0.0559 & 0.3802 & 0.2278 & 0.1531 & 3.8368 & \textcolor{blue}{0.1971} \\
			& GradCAM++ & 0.1334 & 0.4303 & 0.1303 & 0.3452 & 0.3475 & 0.1815 & 3.7896 & 0.2063 \\
			& ODAM & 0.2274 & 0.3219 & 0.3363 & 0.2254 & 0.8290 & \textcolor{blue}{0.3537} & \textcolor{blue}{2.8318} & 0.2008 \\
			& D-RISE & \textcolor{blue}{0.2799} & \textcolor{blue}{0.2914} & \textcolor{blue}{0.3641} & 0.1936 & \textcolor{blue}{0.8834} & 0.1122 & 4.0916 & 43.3966 \\
			& VPS & 0.2656 & \textcolor{red}{0.2871} & 0.2984 & \textcolor{red}{0.1658} & 0.4412 & 0.2249 & 3.7928 & 105.9951 \\
			& EIVE (ours) & \textcolor{red}{0.3126} & 0.3073 & \textcolor{red}{0.4569} & \textcolor{blue}{0.1925} & \textcolor{red}{0.9969} & \textcolor{red}{0.9485} & \textcolor{red}{0.5034} & \textcolor{red}{0.0774} \\
			\hline
		\end{tabular}
	}
\end{table*}
\begin{figure*}[!t]
	\centering
	\includegraphics[width=\textwidth]{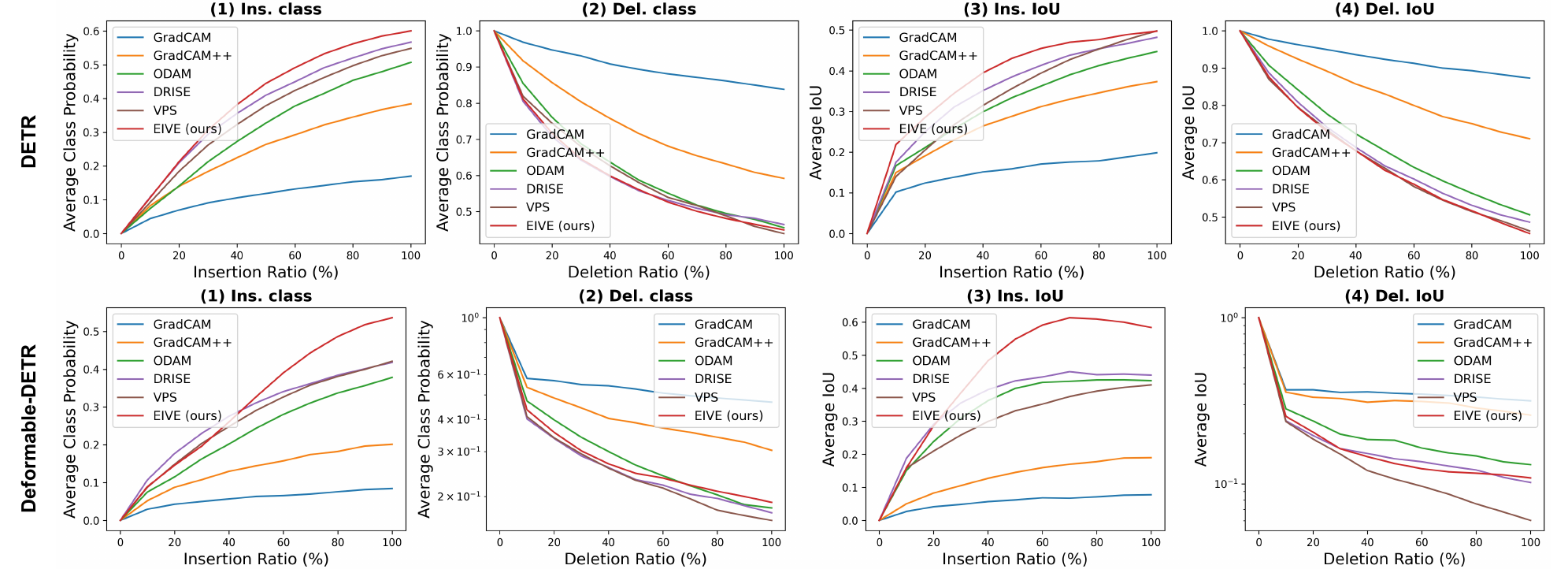}
	\caption{Average Insertion and Deletion Curves for Faithfulness Evaluation on the ExDark Dataset.}
	\label{fig:Insert-delete-curves-exdark}
\end{figure*}

\subsection{Evaluation Metrics}
\label{sec:evaluation_metrics}
We evaluate instance-level visual explanations in object detection from three perspectives: faithfulness, spatial localization, and computational efficiency.

For explanation faithfulness, we adopt Insertion and Deletion experiments~\cite{ref_D-RISE}. These metrics progressively insert or remove image regions according to their saliency scores and measure the resulting changes in detection outputs. Insertion (class) and Deletion (class) evaluate the variation of predicted class confidence, while Insertion (IoU) and Deletion (IoU) evaluate the variation of localization quality using the IoU between the predicted box and the ground-truth box. A faithful explanation should lead to a rapid decrease in confidence and IoU when salient regions are removed, and a consistent increase when they are inserted.

To evaluate spatial localization ability, we use the Pointing Game (PG) metric~\cite{ref_PG}, which checks whether the maximum activation point of a saliency map falls inside the annotated target region. Since PG only considers the strongest response, we further adopt Energy Pointing Game (Energy PG)~\cite{ref_ODAM}, which measures the proportion of saliency energy inside the ground-truth region. We also report Compactness (Comp.)~\cite{ref_ODAM} to measure the spatial spread of saliency around the maximum activation point, where a smaller value indicates a more compact explanation.

Furthermore, to assess the practical efficiency of explanation methods, we report the average explanation time per object instance (Time per Object, in seconds). This metric measures the additional computational overhead required to generate a saliency map for each target instance, excluding the standard detector inference used to obtain the original detection results. Any extra operations introduced by the explainer, such as gradient backpropagation or repeated perturbed inference, are included in this time.

\begin{table*}[!t]
\caption{Quantitative comparison results of different visual explanation methods on the Cityscapes dataset. Time (s) denotes the average explanation time required for a single target instance. \textcolor{red}{Red} and \textcolor{blue}{blue} indicate the best and second-best results, respectively.}
\label{tab:cityscapes_explanation_comparison}
\centering
\renewcommand{\arraystretch}{1.1}
\resizebox{\textwidth}{!}{
	\begin{tabular}{@{\hspace{0.6em}}
			>{\arraybackslash}m{2.0cm}
			>{\arraybackslash}m{1.5cm} |
			>{\centering\arraybackslash}m{1.5cm}
			>{\centering\arraybackslash}m{1.5cm}
			>{\centering\arraybackslash}m{1.3cm}
			>{\centering\arraybackslash}m{1.3cm} |
			>{\centering\arraybackslash}m{1.7cm}
			>{\centering\arraybackslash}m{1.6cm}
			>{\centering\arraybackslash}m{1.3cm} |
			>{\centering\arraybackslash}m{1.8cm}
			@{\hspace{0.6em}}}
		\hline
		\textbf{Detectors} & \textbf{Methods} &
		\textbf{Ins.\ Class $\uparrow$} &
		\textbf{Del.\ Class $\downarrow$} &
		\textbf{Ins.\ IoU $\uparrow$} &
		\textbf{Del.\ IoU $\downarrow$} &
		\textbf{Point Game $\uparrow$} &
		\textbf{Energy PG $\uparrow$} &
		\textbf{Comp.\ $\downarrow$} &
		\textbf{Time (s) $\downarrow$} \\
		\hline
		\multirow{6}{*}{{DETR}}
		& GradCAM & 0.0018 & 0.9974 & 0.0114 & 0.9808 & 0.0493 & 0.0207 & 16.6288 & \textcolor{blue}{0.4378} \\
		& GradCAM++ & 0.0045 & 0.9866 & 0.0158 & 0.9726 & 0.0575 & 0.0271 & 16.0615 & 0.4403 \\
		& ODAM & 0.0071 & \textcolor{blue}{0.8762} & \textcolor{blue}{0.0290} & \textcolor{blue}{0.7943} & \textcolor{blue}{0.5785} & \textcolor{blue}{0.1224} & \textcolor{blue}{13.2816} & 0.4511 \\
		& D-RISE & 0.0029 & 0.8979 & 0.0175 & 0.8033 & 0.3997 & 0.0096 & 17.3673 & 6.6986 \\
		& VPS & \textcolor{blue}{0.0108} & 0.9595 & 0.0248 & 0.8979 & 0.0602 & 0.0333 & 18.0412 & 131.6346 \\
		& EIVE (ours) & \textcolor{red}{0.0109} & \textcolor{red}{0.8382} & \textcolor{red}{0.0398} & \textcolor{red}{0.7287} & \textcolor{red}{0.7797} & \textcolor{red}{0.6380} & \textcolor{red}{0.8620} & \textcolor{red}{0.3381} \\
		\hline
		\multirow{6}{*}{{Deformable-DETR}}
		& GradCAM & 0.0062 & 0.2123 & 0.0057 & 0.1359 & 0.0189 & 0.0107 & 18.2710 & \textcolor{blue}{0.5287} \\
		& GradCAM++ & 0.0084 & 0.2166 & 0.0073 & 0.1387 & 0.0269 & 0.0157 & 18.3150 & 0.5353 \\
		& ODAM & 0.1394 & \textcolor{blue}{0.1527} & \textcolor{blue}{0.3315} & 0.0846 & \textcolor{blue}{0.6480} & \textcolor{blue}{0.1327} & \textcolor{blue}{10.1142} & 0.5366 \\
		& D-RISE & \textcolor{blue}{0.1707} & \textcolor{red}{0.1422} & 0.2732 & \textcolor{red}{0.0802} & 0.4578 & 0.0105 & 16.0226 & 16.0041 \\
		& VPS & 0.0874 & 0.1630 & 0.0931 & 0.1011 & 0.0962 & 0.0495 & 15.9661 & 238.4136 \\
		& EIVE (ours) & \textcolor{red}{0.2349} & 0.1558 & \textcolor{red}{0.5589} & \textcolor{blue}{0.0843} & \textcolor{red}{0.9976} & \textcolor{red}{0.8591} & \textcolor{red}{0.5878} & \textcolor{red}{0.3677} \\
		\hline
	\end{tabular}
}
\end{table*}
\begin{figure*}[!t]
\centering
\includegraphics[width=\textwidth]{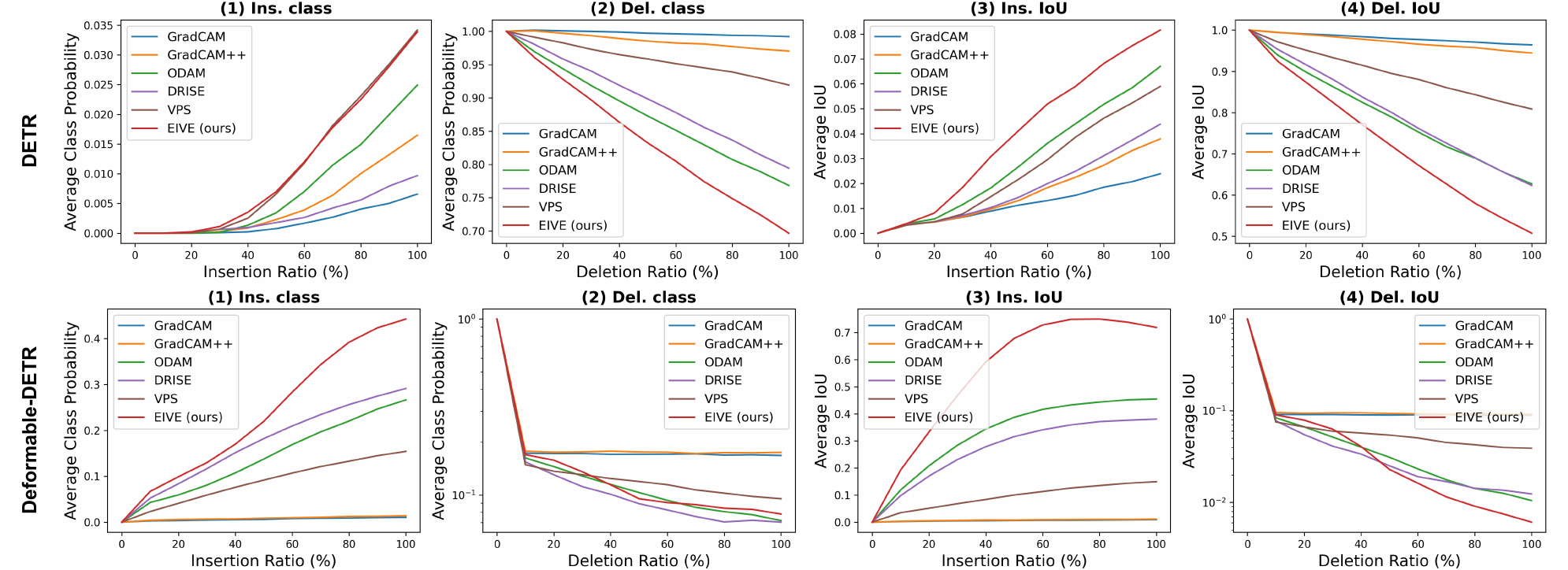}
\caption{Average Insertion and Deletion Curves for Faithfulness Evaluation on the Cityscapes Dataset.}
\label{fig:Insert-delete-curves-cityscapes}
\end{figure*}

\subsection{Quantitative Evaluation of Visual Explanations}
We conduct a quantitative evaluation of the proposed EIVE method against several representative baselines on three widely used benchmarks, namely COCO, ExDark, and Cityscapes. The evaluation follows the metrics described in Section~\ref{sec:evaluation_metrics}. For each dataset, we interpret only correctly detected instances, where the predicted category matches the ground-truth category and both the classification confidence and IoU with the matched ground-truth box are greater than 0.5. This protocol ensures that the explanation metrics are computed on valid detection results. For completeness, the detection performance of the detectors used for explanation analysis is summarized in Table~\ref{tab:detector_performance}. Specifically, all detectors are initialized from the COCO-pretrained checkpoints provided by MMDetection. For COCO, these checkpoints are directly used for explanation analysis, while for ExDark and Cityscapes, the corresponding detectors are further fine-tuned on each dataset. Based on the resulting fixed detectors, the quantitative comparison results are reported in Table~\ref{tab:coco_explanation_comparison}, Table~\ref{tab:exdark_explanation_comparison}, and Table~\ref{tab:cityscapes_explanation_comparison}. We evaluate five DETR-family detectors on COCO, while reporting representative results on DETR and Deformable-DETR for ExDark and Cityscapes.

\subsubsection{Faithfulness Evaluation} 
To verify whether the generated explanations faithfully reflect the decision basis of object detectors, we evaluate EIVE on COCO, ExDark, and Cityscapes using Insertion and Deletion metrics from both class confidence and bounding box IoU perspectives. Overall, EIVE achieves superior faithfulness across different DETR-like architectures.

On COCO, EIVE obtains the best Ins. Class and Ins. IoU scores for most detectors, with more pronounced advantages on stronger models. For example, on Deformable-DETR, EIVE improves Ins. Class to 0.3872 and Ins. IoU to 0.4765, clearly outperforming VPS, whose Ins. Class score is 0.3102 and Ins. IoU score is 0.2549. On DINO, EIVE achieves the best Ins. Class score of 0.3179, and also attains the highest Ins. IoU score of 0.4459. Under the Deletion metric, EIVE produces lower or competitive Del. Class and Del. IoU values, indicating that removing the identified regions leads to faster performance degradation in most settings. In comparison, Grad-CAM and Grad-CAM++ generally show weaker performance across all four metrics, while D-RISE and VPS provide only partial gains on some settings. 

On ExDark, EIVE maintains strong faithfulness under low-light conditions. For DETR, compared with the strongest baseline, EIVE improves Ins. Class from 0.3669 to 0.3924 and Ins. IoU from 0.3488 to 0.3815. Meanwhile, EIVE also achieves the best deletion results. For Deformable-DETR, EIVE further improves Ins. Class from 0.2799 to 0.3126 and Ins. IoU from 0.3641 to 0.4569. However, VPS obtains better deletion scores. This may be because perturbation-based methods such as VPS tend to activate broader object-related and contextual regions. When these regions are removed, more visual evidence is suppressed simultaneously, which can be favorable for deletion metrics. In contrast, EIVE produces more compact explanations, yet still achieves better insertion performance, indicating that its activated regions are more concentrated while being more effective in recovering detector responses.

On Cityscapes, EIVE also demonstrates strong faithfulness. For DETR, compared with the strongest baseline, EIVE slightly improves Ins. Class from 0.0108 to 0.0109 and increases Ins. IoU from 0.0290 to 0.0398. Meanwhile, EIVE achieves the best deletion results, reducing Del. Class from 0.8762 to 0.8382 and Del. IoU from 0.7943 to 0.7287. For Deformable-DETR, EIVE further improves Ins. Class from 0.1707 to 0.2349 and substantially increases Ins. IoU from 0.3315 to 0.5589, suggesting that the highlighted regions contain more sufficient localization-related evidence. For the deletion metrics, a similar trend to ExDark is observed, where the perturbation-based D-RISE obtains slightly better results. Nevertheless, EIVE still achieves competitive deletion performance.

The average insertion and deletion curves on COCO, ExDark, and Cityscapes, shown in Fig.~\ref{fig:Insert-delete-curves-coco}, Fig.~\ref{fig:Insert-delete-curves-exdark}, and Fig.~\ref{fig:Insert-delete-curves-cityscapes}, further support these results. EIVE consistently exhibits faster increases during insertion and more pronounced drops during deletion, indicating that the highlighted regions more precisely capture the evidence that drives detection outcomes. 

\subsubsection{Spatial Localization Evaluation} 
We further evaluate the spatial localization capability of saliency maps using the Pointing Game, Energy-based Pointing Game (Energy PG), and Compactness metrics. 

On COCO, EIVE shows clear advantages across different detectors. On DETR, EIVE achieves a Pointing Game score of 0.8475 and an Energy PG of 0.7736, both substantially higher than those of ODAM, whose Pointing Game score is 0.5976. On Deformable-DETR, the Pointing Game score of EIVE reaches 0.9956, while Energy PG increases to 0.9163, indicating very strong spatial alignment. At the same time, EIVE consistently produces much lower Compactness values than competing methods. For instance, on Deformable-DETR, its Compactness is only 0.5115, whereas most competing methods exceed 4, showing that EIVE generates more compact and concentrated saliency maps. 

On ExDark, the spatial localization advantage of EIVE becomes even more evident. On DETR and Deformable-DETR, the Pointing Game scores reach 0.9437 and 0.9969, respectively. EIVE also achieves the highest Energy PG values across detectors and maintains the lowest Compactness values.

On Cityscapes, EIVE likewise demonstrates strong localization capability. On DETR, EIVE achieves a Pointing Game score of 0.7797 and an Energy PG of 0.6380, both clearly higher than those of ODAM, while reducing Compactness to 0.8620. On Deformable-DETR, the Pointing Game reaches 0.9976, Energy PG increases to 0.8591, and Compactness drops to 0.5878, indicating highly accurate and compact spatial alignment.

\begin{figure*}[!t]
	\centering
	\includegraphics[width=\textwidth]{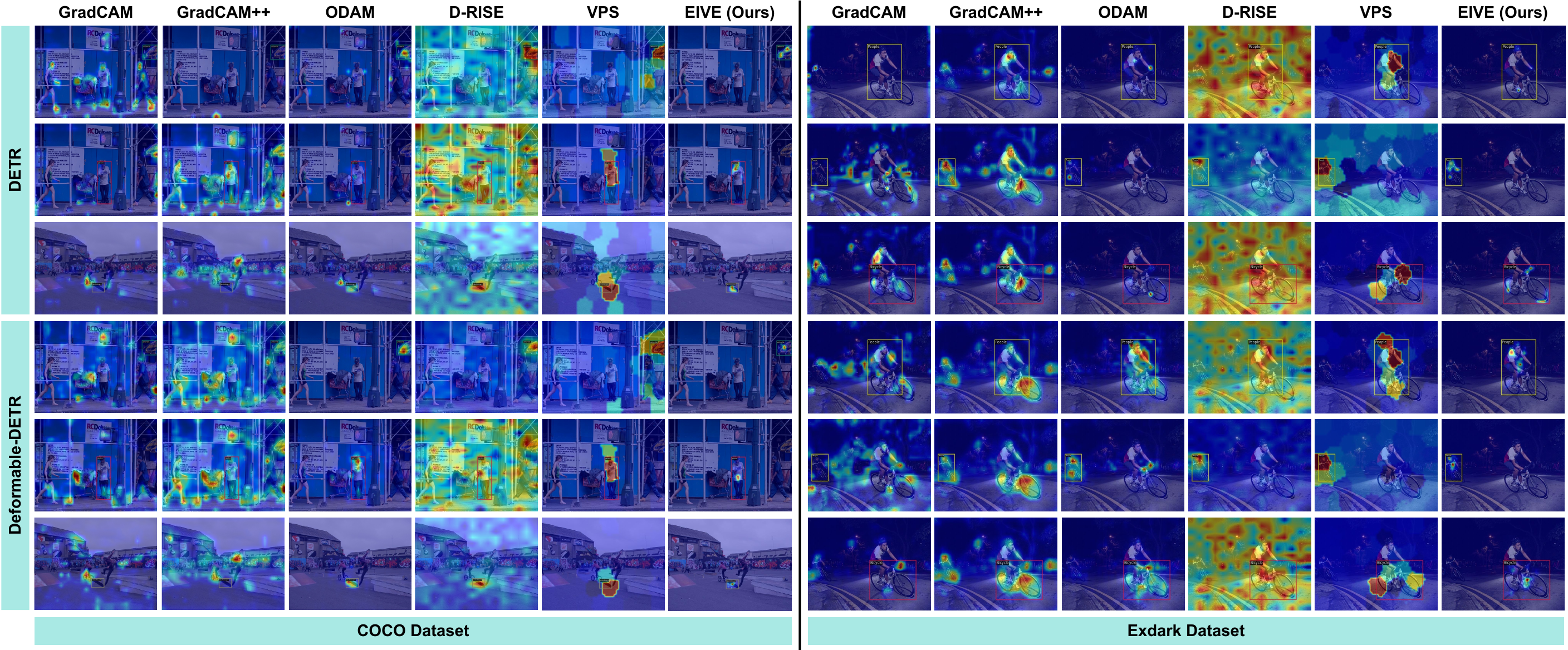}
	\caption{Qualitative comparison of EIVE and state-of-the-art explanation methods on the COCO and ExDark datasets. The left part shows the results on the COCO dataset, and the right part shows the results on the ExDark dataset.}
	\label{fig:qual_coco_exdark}
\end{figure*}

\subsubsection{Computational Efficiency Analysis} 
In addition to explanation quality, computational efficiency is critical for practical use. We therefore measure the average explanation time per object instance. 

On COCO, the computational overhead of EIVE remains stable between 0.0440 seconds and 0.0776 seconds, which is substantially lower than that of perturbation-based methods such as D-RISE and VPS. Specifically, D-RISE typically requires 17 to 55 seconds, while VPS exceeds 60 seconds. Compared with these methods, EIVE achieves speedups of several hundred to more than one thousand times. Even relative to gradient-based methods such as GradCAM and ODAM, EIVE still maintains an approximately threefold to fourfold speed advantage. 

A similar trend is observed on ExDark. EIVE requires only 0.0472 seconds and 0.0774 seconds per instance with DETR and Deformable-DETR, respectively, whereas D-RISE and VPS still incur explanation times on the order of tens of seconds.

On Cityscapes, EIVE maintains the same favorable efficiency profile. The average explanation time per target instance ranges from 0.3381 seconds to 0.3677 seconds across different detectors, which is consistently lower than that of GradCAM, GradCAM++, and ODAM, and dramatically lower than that of perturbation-based methods. In particular, D-RISE requires about 6.70 to 16.00 seconds per instance, while VPS further increases to 131.63 to 238.41 seconds. Although the image-level computation strategy of D-RISE can partially amortize its per-instance cost in multi-object scenes, EIVE still maintains a substantial runtime advantage over D-RISE.

Overall, EIVE achieves substantial improvements in faithfulness and spatial localization while maintaining very high computational efficiency, thereby striking a more favorable balance between explanation quality and runtime cost.

\begin{figure*}[!t]
	\centering
	\includegraphics[width=\textwidth]{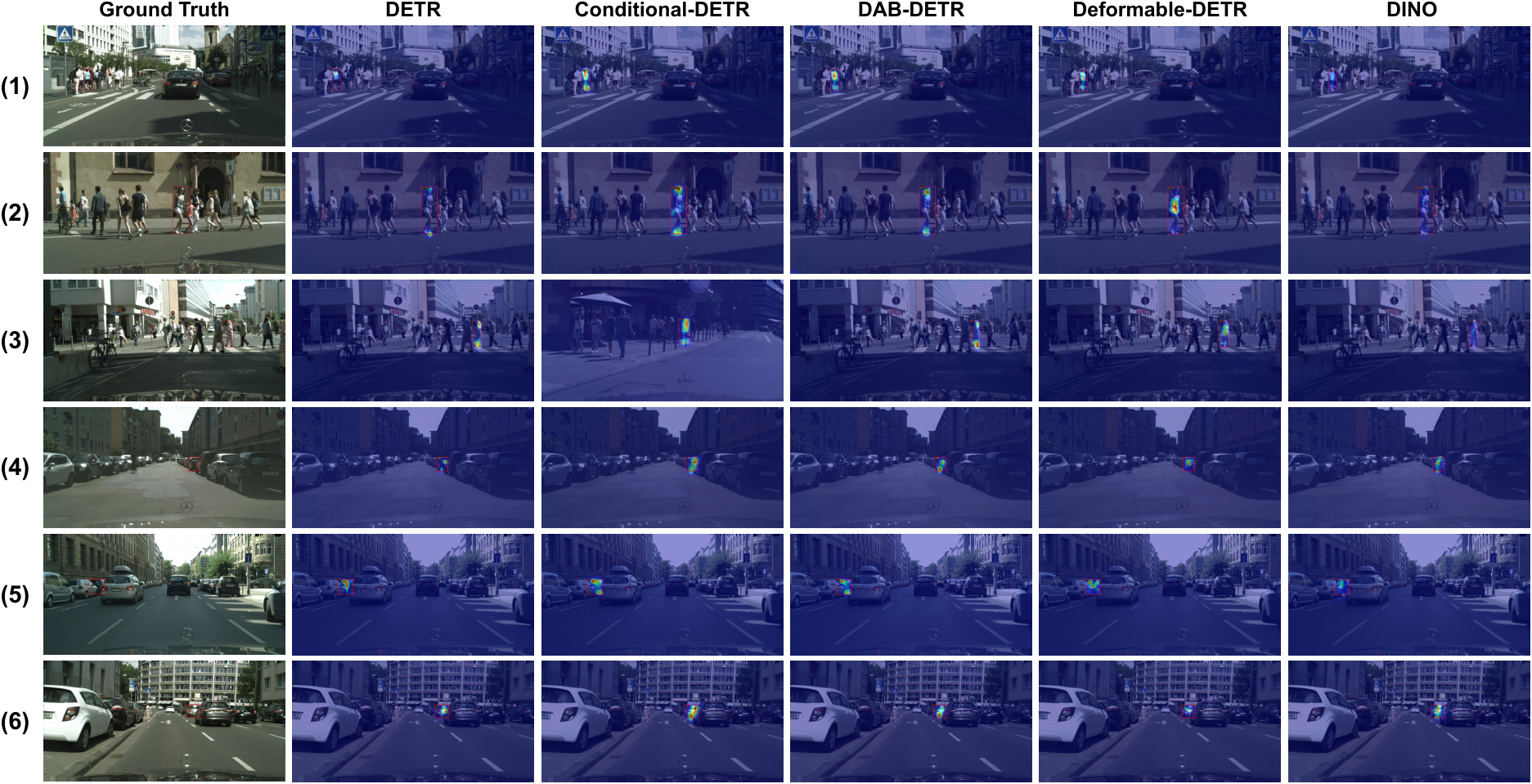}
	\caption{Instance-level explanation results of different Detection Transformer detectors in dense scenes on the Cityscapes dataset. Samples (1)--(3) show the dense instance explanation results for the ``person'' class, while samples (4)--(6) show the dense instance explanation results for the ``car'' class.}
	\label{fig:instance_dense}
\end{figure*}

\begin{figure}[t]
	\centering
	\includegraphics[width=\columnwidth]{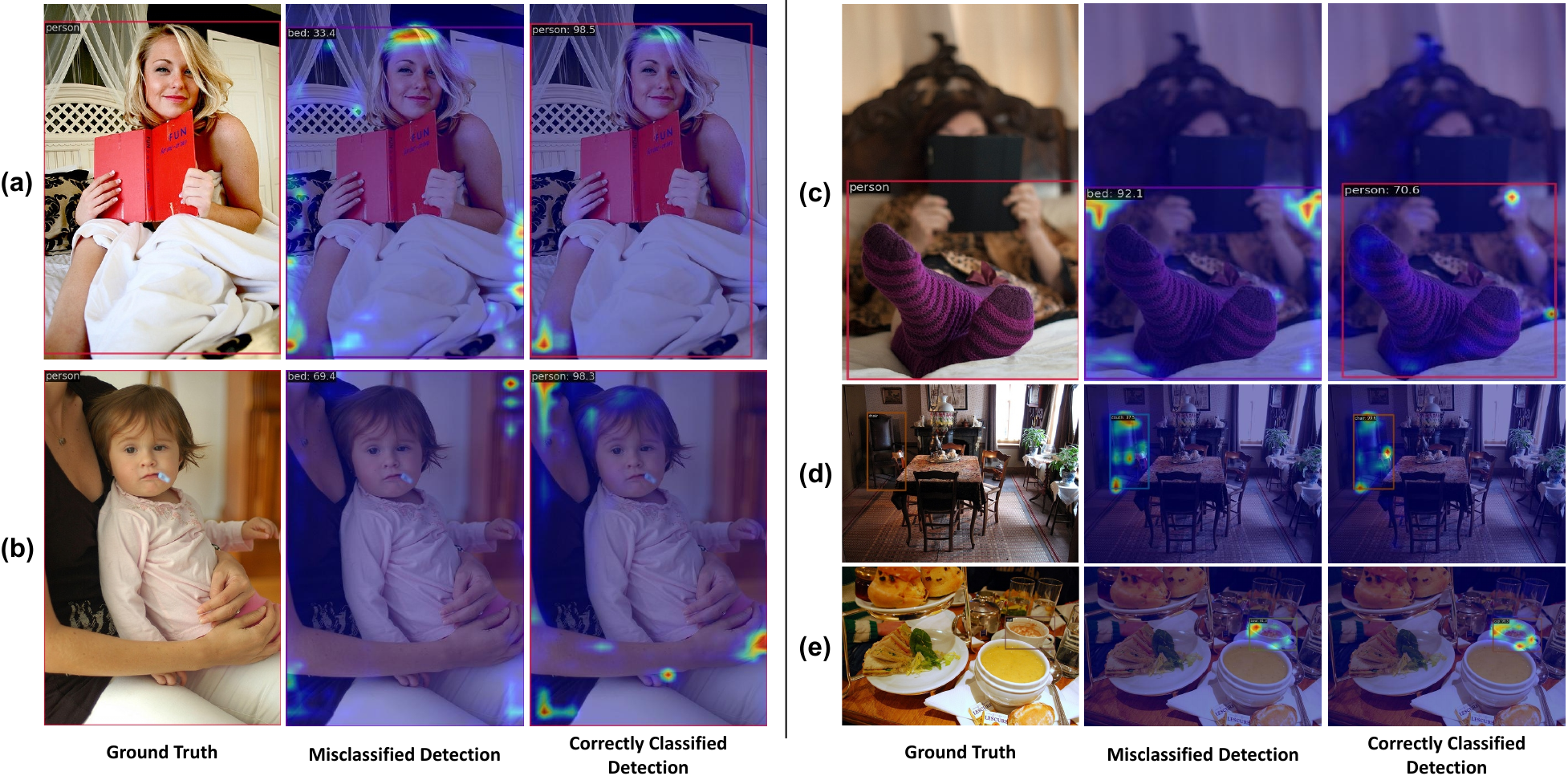}
	\caption{Visualization of typical error modes of the pretrained DETR detector on the COCO dataset.}
	\label{fig:error_Analyzing}
\end{figure}

\subsection{Qualitative Evaluation of Visual Explanations}
\subsubsection{Comparison with State-of-the-Art Explanation Methods}
We further conducted a qualitative visual comparison between EIVE and several mainstream explanation methods on the COCO and ExDark datasets, in order to intuitively evaluate their performance in terms of target-region focus, background-noise suppression, and explanation stability in multi-instance scenes.

For the COCO dataset shown in the left part of Fig.~\ref{fig:qual_coco_exdark}, GradCAM and GradCAM++ generally suffer from dispersed responses in complex backgrounds and multi-instance scenes, where the heatmaps often cover large background areas and fail to concentrate on the interior of target objects. ODAM enhances target-related regions to some extent, but still exhibits local drift in scenes with multiple persons or significant scale variations. Although D-RISE produces relatively continuous response distributions, it introduces strong overall noise, activating large irrelevant regions. VPS, by comparison, provides clearer target responses, but still shows a certain degree of co-activation in complex backgrounds. In contrast, EIVE demonstrates stronger discriminative focus on both detectors. Its heatmaps are mainly concentrated on key structural regions of the predicted targets, with significantly reduced background interference. In multi-object scenes, the response boundaries between different instances are clearer.

For the ExDark dataset shown in the right part of Fig.~\ref{fig:qual_coco_exdark}, GradCAM and GradCAM++ are more easily affected by illumination changes, and their heatmaps tend to focus on locally bright regions rather than the true object body. Under low-light conditions, D-RISE exhibits more severe noise diffusion, resulting in weaker discrimination between targets and backgrounds. Although VPS can cover the overall target region, it still produces block-like false activations in dark backgrounds. In comparison, EIVE remains stable under low-light and high-interference conditions. Its heatmaps are mainly concentrated on the target body and its key structures, while dark background regions show almost no obvious false activation.

Overall, whether in regular complex scenes or in low-illumination environments with large domain shifts, EIVE consistently exhibits superior target focus and lower background interference. These qualitative results are consistent with the quantitative faithfulness evaluation presented earlier, further validating the effectiveness and robustness of the proposed method for visual explanation in object detection.

\subsubsection{Instance-Level Explanations under Dense Scenes}
To further evaluate the instance-level discriminative capability of the proposed method in complex and dense urban scenes, we selected representative samples from the Cityscapes dataset and visualized the instance-level explanations produced by five Detection Transformer detectors, namely DETR, Conditional-DETR, DAB-DETR, Deformable-DETR, and DINO. The selected samples contain crowded pedestrians and closely adjacent vehicles. The results are shown in Fig.~\ref{fig:instance_dense}.

Across different object categories, the explanation maps exhibit consistent semantic alignment with the corresponding targets. For the \textit{person} class shown in samples (1)--(3), the explanation maps can accurately highlight the target pedestrians in crowded scenes and maintain relatively clear boundaries between adjacent instances. For the \textit{car} class shown in samples (4)--(6), the explanations are also highly concentrated on the target vehicles, with limited interference from surrounding objects and background regions.

Overall, the proposed method produces clear and discriminative instance-level explanation maps for dense urban scenes. Even when multiple objects appear in close proximity, the explanations still concentrate well on the target instances and exhibit good spatial separation. This indicates that the proposed method has strong applicability in challenging scenarios with dense layouts and occlusion.

\begin{table*}[t]
	\centering
	\caption{Quantitative results on COCO dataset comparing baseline training and the proposed AAJTS across different detectors and input sizes. ``\xmark'' and ``\cmark'' indicate without and with AAJTS, respectively. \textcolor{red}{Red} indicates the best results.}
	\label{tab:aajts_coco}
	\renewcommand{\arraystretch}{1.1}
	\setlength{\tabcolsep}{2.6pt}
	\resizebox{\textwidth}{!}{
		\begin{tabular}{l|c|c|cccccc|ccccc}
			\hline
			\textbf{Detectors} & \textbf{Size} & \textbf{AAJTS} & \textbf{mAP} & \textbf{mAP$_{50}$} & \textbf{mAP$_{75}$} & \textbf{mAP$_s$} & \textbf{mAP$_m$} & \textbf{mAP$_l$} & \textbf{Ins. Class}$\uparrow$ & \textbf{Ins. IoU}$\uparrow$ & \textbf{Point Game}$\uparrow$ & \textbf{Energy PG}$\uparrow$ & \textbf{Comp.}$\downarrow$ \\
			\hline
			\multirow{4}{*}{DAB-DETR} 
			& \multirow{2}{*}{Fix 512} & \xmark & 0.357 & 0.557 & \textcolor{red}{0.372} & \textcolor{red}{0.131} & 0.378 & 0.581 & 0.1826 & 0.0261 & 0.8359 & 0.7362 & 0.8373 \\
			& & \cmark & \textcolor{red}{0.359} & \textcolor{red}{0.558} & 0.370 & 0.130 & \textcolor{red}{0.384} & \textcolor{red}{0.588} & \textcolor{red}{0.2078} & \textcolor{red}{0.0305} & \textcolor{red}{0.8853} & \textcolor{red}{0.8376} & \textcolor{red}{0.6009} \\
			\cline{2-14}
			& \multirow{2}{*}{Fix 800} & \xmark & 0.404 & 0.607 & 0.428 & \textcolor{red}{0.182} & 0.441 & 0.605 & 0.1758 & 0.0297 & 0.8896 & 0.8013 & 0.8123 \\
			& & \cmark & \textcolor{red}{0.405} & 0.607 & \textcolor{red}{0.431} & 0.178 & 0.441 & \textcolor{red}{0.612} & \textcolor{red}{0.2244} & \textcolor{red}{0.0399} & \textcolor{red}{0.9231} & \textcolor{red}{0.8811} & \textcolor{red}{0.6106} \\
			\cline{1-14}
			\multirow{4}{*}{Deformable-DETR} 
			& \multirow{2}{*}{Fix 512} & \xmark & 0.356 & 0.548 & 0.379 & 0.161 & 0.380 & 0.535 & 0.3363 & 0.4581 & 0.9945 & 0.7262 & 0.7879 \\
			& & \cmark & \textcolor{red}{0.371} & \textcolor{red}{0.557} & \textcolor{red}{0.395} & \textcolor{red}{0.177} & \textcolor{red}{0.393} & \textcolor{red}{0.547} & \textcolor{red}{0.3421} & \textcolor{red}{0.5050} & \textcolor{red}{0.9951} & \textcolor{red}{0.9225} & \textcolor{red}{0.4465} \\
			\cline{2-14}
			& \multirow{2}{*}{Fix 800} & \xmark & 0.382 & 0.576 & 0.408 & 0.210 & 0.414 & 0.535 & 0.3270 & 0.4461 & 0.9912 & 0.7948 & 0.6869 \\
			& & \cmark & \textcolor{red}{0.396} & \textcolor{red}{0.588} & \textcolor{red}{0.426} & \textcolor{red}{0.212} & \textcolor{red}{0.427} & \textcolor{red}{0.554} & \textcolor{red}{0.3571} & \textcolor{red}{0.4837} & \textcolor{red}{0.9935} & \textcolor{red}{0.9381} & \textcolor{red}{0.4594} \\
			\hline
		\end{tabular}
	}
\end{table*}
\begin{table*}[t]
	\centering
	\caption{Quantitative results on ExDark dataset comparing baseline training and the proposed AAJTS across different detectors and input sizes. ``\xmark'' and ``\cmark'' indicate without and with AAJTS, respectively. \textcolor{red}{Red} indicates the best results.}
	\label{tab:aajts_exdark}
	\renewcommand{\arraystretch}{1.1}
	\setlength{\tabcolsep}{2.6pt}
	\resizebox{\textwidth}{!}{
		\begin{tabular}{l|c|c|cccccc|ccccc}
			\hline
			\textbf{Detectors} & \textbf{Size} & \textbf{AAJTS} & \textbf{mAP} & \textbf{mAP$_{50}$} & \textbf{mAP$_{75}$} & \textbf{mAP$_s$} & \textbf{mAP$_m$} & \textbf{mAP$_l$} & \textbf{Ins. Class}$\uparrow$ & \textbf{Ins. IoU}$\uparrow$ & \textbf{Point Game}$\uparrow$ & \textbf{Energy PG}$\uparrow$ & \textbf{Comp.}$\downarrow$ \\
			\hline
			\multirow{4}{*}{DAB-DETR} 
			& \multirow{2}{*}{Fix 512} & \xmark & 0.336 & 0.636 & 0.323 & \textcolor{red}{0.055} & 0.174 & 0.434 & 0.2588 & 0.1913 & 0.9688 & 0.5998 & 1.4602 \\
			& & \cmark & \textcolor{red}{0.351} & \textcolor{red}{0.646} & \textcolor{red}{0.345} & 0.048 & \textcolor{red}{0.192} & \textcolor{red}{0.449} & \textcolor{red}{0.3061} & \textcolor{red}{0.2235} & \textcolor{red}{0.9817} & \textcolor{red}{0.9318} & \textcolor{red}{0.4056} \\
			\cline{2-14}
			& \multirow{2}{*}{Fix 800} & \xmark & 0.370 & 0.683 & 0.353 & \textcolor{red}{0.087} & 0.235 & 0.455 & 0.2399 & 0.1653 & 0.9843 & 0.8903 & 0.6431 \\
			& & \cmark & \textcolor{red}{0.381} & \textcolor{red}{0.696} & \textcolor{red}{0.380} & 0.052 & \textcolor{red}{0.244} & \textcolor{red}{0.465} & \textcolor{red}{0.3273} & \textcolor{red}{0.2316} & \textcolor{red}{0.9896} & \textcolor{red}{0.9506} & \textcolor{red}{0.4499} \\
			\cline{1-14}
			\multirow{4}{*}{Deformable-DETR} 
			& \multirow{2}{*}{Fix 512} & \xmark & 0.302 & 0.541 & 0.303 & 0.041 & \textcolor{red}{0.174} & 0.379 & 0.2857 & 0.5159 & \textcolor{red}{0.9962} & 0.8520 & 0.6093 \\
			& & \cmark & \textcolor{red}{0.309} & \textcolor{red}{0.554} & \textcolor{red}{0.315} & 0.041 & 0.172 & \textcolor{red}{0.391} & \textcolor{red}{0.3384} & \textcolor{red}{0.5174} & 0.9952 & \textcolor{red}{0.9216} & \textcolor{red}{0.5366} \\
			\cline{2-14}
			& \multirow{2}{*}{Fix 800} & \xmark & 0.310 & 0.540 & 0.323 & 0.049 & 0.207 & 0.379 & 0.2836 & 0.4619 & 0.9779 & 0.8591 & 0.6936 \\
			& & \cmark & \textcolor{red}{0.319} & \textcolor{red}{0.557} & \textcolor{red}{0.335} & \textcolor{red}{0.070} & \textcolor{red}{0.213} & \textcolor{red}{0.385} & \textcolor{red}{0.3029} & \textcolor{red}{0.5152} & \textcolor{red}{0.9935} & \textcolor{red}{0.9289} & \textcolor{red}{0.5487} \\
			\hline
		\end{tabular}
	}
\end{table*}

\subsubsection{Analyzing Error Modes of the Detector}
Next, we analyze classification errors of a pretrained DETR detector on the COCO dataset. For each ground-truth object, we select the prediction with the largest IoU. If its IoU is greater than 0.9 but its predicted class differs from the ground-truth class, the corresponding object query is regarded as a classification-error sample. For this analysis, we directly visualize the last-decoder-layer cross-attention map of the erroneous query without applying CLHCF, so that the displayed response reflects the final prediction layer. The results are shown in Fig.~\ref{fig:error_Analyzing}.

As shown in \hyperref[fig:error_Analyzing]{Fig.~\ref*{fig:error_Analyzing}(a)}, when the model mainly focuses on the human body region, it can correctly classify the target as ``person''. However, when the attention is also dispersed over surrounding objects such as a quilt, pillow, and wooden board, the model is easily distracted by such contextual information and thus incorrectly classifies the target as ``bed''. Similar phenomena can also be observed in \hyperref[fig:error_Analyzing]{Fig.~\ref*{fig:error_Analyzing}(b)} and \hyperref[fig:error_Analyzing]{Fig.~\ref*{fig:error_Analyzing}(c)}. When the model attends to key human regions such as the arms, it can correctly recognize the target as ``person''. However, when more attention is allocated to background objects such as pillows or wooden boards, the model tends to misclassify the target as ``bed''.

As shown in \hyperref[fig:error_Analyzing]{Fig.~\ref*{fig:error_Analyzing}(d)}, when the model focuses on structural features such as the wooden armrest of the chair, it can correctly recognize the target as ``chair''. However, when the model also attends to the central cushion region, whose appearance is somewhat similar to that of a sofa, it is misled and thus incorrectly classifies the target as ``couch''.

As shown in \hyperref[fig:error_Analyzing]{Fig.~\ref*{fig:error_Analyzing}(e)}, when the model captures discriminative features such as the handle of the cup, it can correctly classify the target as ``cup''. If the model fails to attend to this key structure and instead focuses more on the overall container shape, it tends to misclassify the target as ``bowl''.

These examples show that EIVE can effectively reveal the visual evidence underlying the classification errors of the detector. By comparing the explanation regions of correct and incorrect predictions, it becomes clear that misclassification often occurs when the model attends excessively to contextual regions or non-discriminative structures instead of category-specific object parts. This further demonstrates that EIVE is not only useful for interpreting correct detections, but also effective for diagnosing the error patterns of object detectors.

\subsection{Effectiveness of the AAJTS}
\label{sec:eajts}
To evaluate the effectiveness of the proposed AAJTS, we conduct experiments on both the COCO and ExDark datasets using two representative DETR-like detectors, i.e., DAB-DETR and Deformable-DETR, under two fixed input sizes. In addition to standard detection metrics, we also report insertion-based metrics and localization-based metrics to evaluate the interpretability of the generated saliency maps. 

The results on the COCO dataset are shown in Table~\ref{tab:aajts_coco}. For DAB-DETR, the gains in detection performance under both input sizes are limited, with the mAP slightly varying from 0.357 to 0.359 at Fix 512 and from 0.404 to 0.405 at Fix 800. Nevertheless, even when the detection accuracy remains essentially on par with the baseline, AAJTS still improves the interpretability of the saliency maps. Specifically, at Fix 512, Ins. Class rises from 0.1826 to 0.2078, Ins. IoU from 0.0261 to 0.0305, Point Game from 0.8359 to 0.8853, and Energy PG from 0.7362 to 0.8376, while Comp. drops markedly from 0.8373 to 0.6009. Analogous trends are observed at Fix 800. In contrast, for Deformable-DETR, AAJTS yields clear improvements in both detection accuracy and interpretability. The mAP increases from 0.356 to 0.371 at Fix 512 and from 0.382 to 0.396 at Fix 800, with consistent gains in mAP\(_{50}\), mAP\(_{75}\), and the scale-specific metrics. Concurrently, all attribution-related metrics improve notably. For example, at Fix 800, Energy PG rises from 0.7948 to 0.9381 and Comp. decreases from 0.6869 to 0.4594, indicating that the model focuses more accurately and compactly on target regions.

The results on the ExDark dataset are shown in Table~\ref{tab:aajts_exdark}. AAJTS brings even more pronounced improvements under low-light conditions across both detectors and input sizes. For DAB-DETR, the mAP rises from 0.336 to 0.351 at Fix 512 and from 0.370 to 0.381 at Fix 800, accompanied by consistent gains in mAP\(_{50}\), mAP\(_{75}\), and mAP\(_l\). The interpretability metrics also improve substantially. For instance, at Fix 512, Energy PG increases from 0.5998 to 0.9318 and Comp. drops dramatically from 1.4602 to 0.4056. For Deformable-DETR, AAJTS likewise lifts the mAP from 0.302 to 0.309 at Fix 512 and from 0.310 to 0.319 at Fix 800, with consistent gains across most detection and attribution metrics. By introducing attention-aware supervision, AAJTS guides the model to concentrate on target-related features and suppress irrelevant background responses, which is particularly beneficial under such degraded conditions.

\begin{table*}[t]
	\centering
	\caption{
		Ablation study on the number of fused decoder layers $L'$ in the proposed CLHCF module on pretrained DETR detector. \textcolor{red}{Red} and \textcolor{blue}{blue} indicate the best and second-best results, respectively.
	}
	\label{tab:clhcf_layer_ablation}
	\renewcommand{\arraystretch}{1.1}
	\resizebox{\textwidth}{!}{
		\begin{tabular}{@{\hspace{0.2em}}
				>{\arraybackslash}m{1.2cm} |
				>{\centering\arraybackslash}m{1.2cm} |
				>{\centering\arraybackslash}m{1.8cm}
				>{\centering\arraybackslash}m{1.8cm}
				>{\centering\arraybackslash}m{1.8cm}
				>{\centering\arraybackslash}m{1.8cm} |
				>{\centering\arraybackslash}m{1.8cm}
				>{\centering\arraybackslash}m{1.8cm}
				>{\centering\arraybackslash}m{1.5cm}
				@{\hspace{0.2em}}}
			\hline
			\textbf{Datasets} & $L'$ 
			& \textbf{Ins. Class $\uparrow$} 
			& \textbf{Del. Class $\downarrow$} 
			& \textbf{Ins. IoU $\uparrow$} 
			& \textbf{Del. IoU $\downarrow$} 
			& \textbf{Point Game $\uparrow$} 
			& \textbf{Energy PG $\uparrow$} 
			& \textbf{Comp. $\downarrow$} \\
			
			\hline
			\multirow{6}{*}{COCO}
			& 1 & 0.2238 & 0.7399 & \textcolor{blue}{0.2435} & 0.6963 & 0.8217 & 0.7534 & 0.8920 \\
			& 2 & 0.2161 & 0.7510 & 0.2363 & 0.6945 & 0.8259 & 0.7435 & 0.9069 \\
			& 3 & \textcolor{blue}{0.2263} & \textcolor{blue}{0.7333} & 0.2430 & 0.6903 & 0.8311 & 0.7569 & 0.8798 \\
			& 4 & 0.2230 & 0.7375 & 0.2408 & 0.6898 & 0.8338 & 0.7623 & 0.8660 \\
			& 5 & 0.2249 & 0.7358 & 0.2422 & \textcolor{blue}{0.6896} & \textcolor{blue}{0.8394} & \textcolor{blue}{0.7664} & \textcolor{blue}{0.8555} \\
			& 6 & \textcolor{red}{0.2287} & \textcolor{red}{0.7295} & \textcolor{red}{0.2444} & \textcolor{red}{0.6881} & \textcolor{red}{0.8475} & \textcolor{red}{0.7736} & \textcolor{red}{0.8389} \\
			
			\hline
			\multirow{6}{*}{ExDark}
			& 1 & 0.3780 & 0.6111 & 0.3708 & 0.6670 & 0.9103 & 0.8681 & 0.7867 \\
			& 2 & 0.3545 & 0.6241 & 0.3530 & 0.6756 & 0.9187 & 0.8615 & 0.7782 \\
			& 3 & 0.3839 & 0.6063 & 0.3750 & 0.6594 & 0.9248 & 0.8808 & 0.7522 \\
			& 4 & 0.3830 & 0.6060 & 0.3737 & 0.6613 & 0.9321 & 0.8887 & 0.7306 \\
			& 5 & \textcolor{blue}{0.3869} & \textcolor{blue}{0.6046} & \textcolor{blue}{0.3762} & \textcolor{blue}{0.6582} & \textcolor{blue}{0.9377} & \textcolor{blue}{0.8936} & \textcolor{blue}{0.7155} \\
			& 6 & \textcolor{red}{0.3924} & \textcolor{red}{0.6029} & \textcolor{red}{0.3815} & \textcolor{red}{0.6568} & \textcolor{red}{0.9436} & \textcolor{red}{0.9008} & \textcolor{red}{0.7049} \\
			
			\hline
		\end{tabular}
	}
\end{table*}

\begin{figure*}[t]
	\centering
	\includegraphics[width=\textwidth]{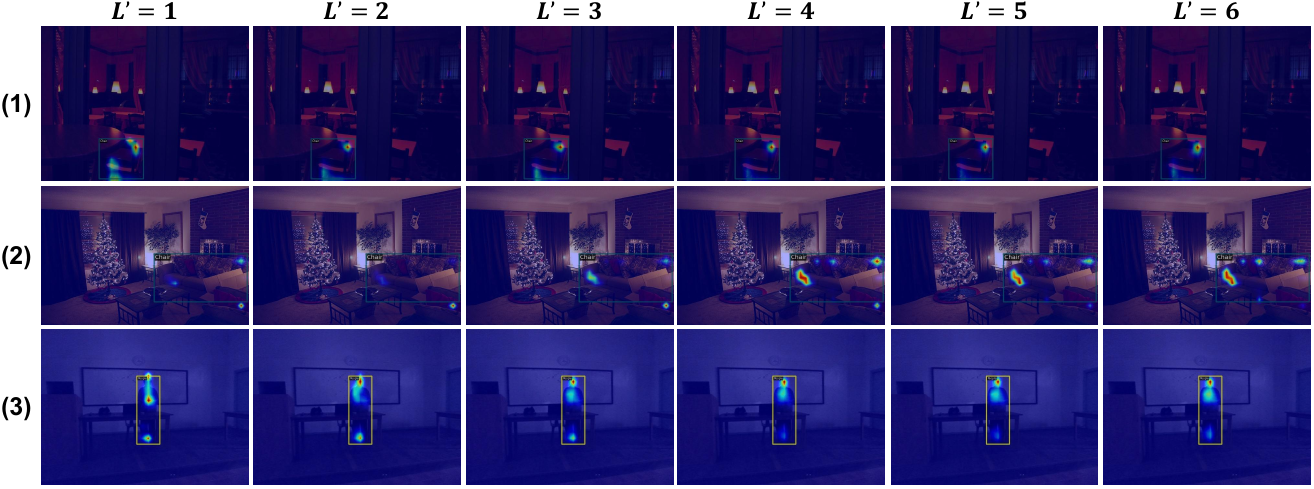}
	\caption{Visualization results of CLHCF with different numbers of fused decoder layers.}
	\label{fig:ablation_layer_visualization}
\end{figure*}

\begin{table}[t]
	\centering
	\caption{
		Ablation study of CLHCF fusion variants on pretrained Deformable DETR. Mean denotes inter-layer averaging, Arith. and Geom. denote arithmetic-only and geometric-only fusion, and Hybrid denotes the full CLHCF strategy. \textcolor{red}{Red} and \textcolor{blue}{blue} indicate the best and second-best results, respectively.
	}
	\label{tab:clhcf_ablation}
	\renewcommand{\arraystretch}{1.1}
	\resizebox{\columnwidth}{!}{
		\begin{tabular}{@{\hspace{0.0em}}
				>{\arraybackslash}m{1.05cm} |
				>{\centering\arraybackslash}m{2.1cm} |
				>{\centering\arraybackslash}m{1.7cm}
				>{\centering\arraybackslash}m{1.6cm}
				>{\centering\arraybackslash}m{1.1cm}
				@{\hspace{0.0em}}}
			\hline
			\textbf{Datasets} & \textbf{CLHCF Variants}
			& \textbf{Point Game $\uparrow$}
			& \textbf{Energy PG $\uparrow$}
			& \textbf{Comp. $\downarrow$} \\
			\hline
			\multirow{4}{*}{COCO}
			& Mean          & 0.9951 & 0.7452 & 0.8339 \\
			& Arith.        & 0.9946 & 0.7420 & 0.8420 \\
			& Geom.         & \textcolor{blue}{0.9955} & \textcolor{blue}{0.8055} & \textcolor{blue}{0.7304} \\
			& Hybrid (Ours) & \textcolor{red}{0.9956} & \textcolor{red}{0.9163} & \textcolor{red}{0.5115} \\
			\hline
			\multirow{4}{*}{ExDark}
			& Mean          & 0.9948 & 0.7987 & 0.7145 \\
			& Arith.        & 0.9965 & 0.7959 & 0.7233 \\
			& Geom.         & \textcolor{blue}{0.9966} & \textcolor{blue}{0.8531} & \textcolor{blue}{0.6312} \\
			& Hybrid (Ours) & \textcolor{red}{0.9969} & \textcolor{red}{0.9485} & \textcolor{red}{0.5034} \\
			\hline
			\multirow{4}{*}{Cityscapes}
			& Mean          & 0.9956 & 0.6247 & 0.9925 \\
			& Arith.        & 0.9961 & 0.6214 & 0.9954 \\
			& Geom.         & \textcolor{blue}{0.9968} & \textcolor{blue}{0.6690} & \textcolor{blue}{0.9100} \\
			& Hybrid (Ours) & \textcolor{red}{0.9976} & \textcolor{red}{0.8591} & \textcolor{red}{0.5878} \\
			\hline
		\end{tabular}
	}
\end{table}
\begin{figure}[t]
	\centering
	\includegraphics[width=\columnwidth]{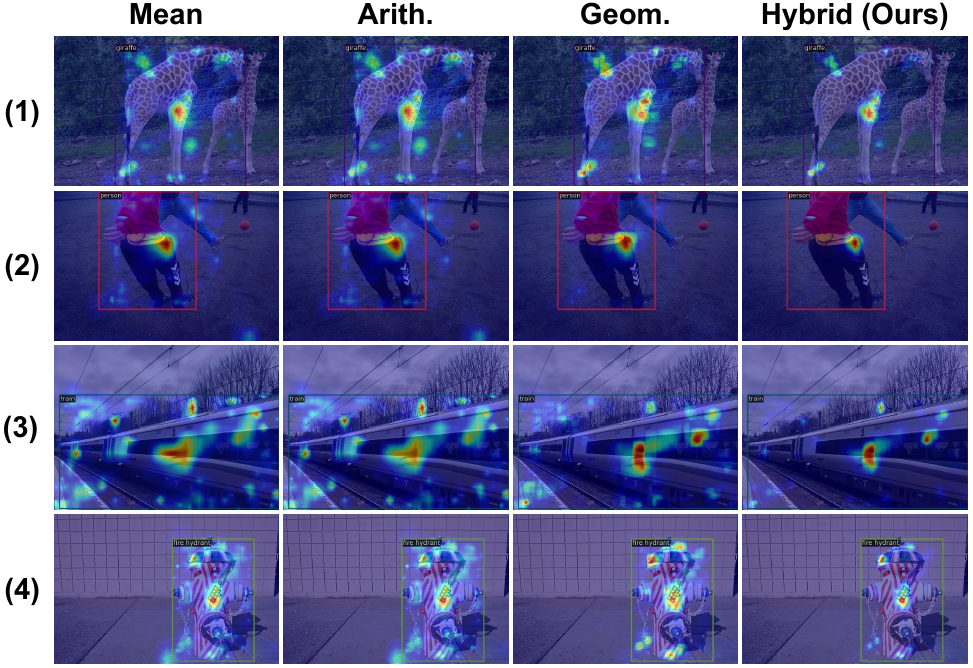}
	\caption{Visualization results of different CLHCF fusion variants.}
	\label{fig:ablation_hybrid_visualization}
\end{figure}

\subsection{Ablation Study}
\label{sec:ablation_clhcf}
We conduct ablation studies to analyze the effectiveness of the proposed Cross-Layer Hybrid Consensus Fusion (CLHCF) module from two perspectives. First, we study the effect of the number of fused decoder layers $L'$ on a pretrained DETR detector, as shown in Table~\ref{tab:clhcf_layer_ablation}. Second, we isolate different fusion variants of CLHCF on a pretrained Deformable DETR detector, as reported in Table~\ref{tab:clhcf_ablation}. These experiments are designed to verify whether cross-layer aggregation is beneficial and whether the proposed hybrid fusion strategy is necessary.

\subsubsection{Effect of the Number of Fused Decoder Layers}
\label{sec:ablation_clhcf_layers}
Table~\ref{tab:clhcf_layer_ablation} shows that using more decoder layers generally improves the quality of the explanation maps. Compared with using only the last decoder layer ($L'=1$), fusing all six layers improves Energy PG by 3.77\% and reduces Comp. by 10.40\% on ExDark. On COCO, the corresponding improvement is 2.68\% in Energy PG and 5.95\% in Comp. The advantage of $L'=6$ also holds when compared with the strongest partial-fusion setting among $L'<6$. These gains indicate that different decoder layers provide complementary spatial evidence, and relying on only the final layer may miss useful attribution cues from earlier layers.

We also observe that the improvement is not strictly monotonic for all intermediate settings. For instance, $L'=2$ performs worse than $L'=1$ on several faithfulness metrics. This observation suggests that the contribution of each decoder layer is not uniformly beneficial for every metric. Therefore, the improvement should not be interpreted as a simple consequence of adding more layers. Instead, the results indicate that cross-layer fusion is useful when sufficient layer-level evidence is aggregated under the proposed query tracking and reliability-weighted fusion scheme.

The qualitative results in Fig.~\ref{fig:ablation_layer_visualization} provide consistent visual evidence for this observation. When only a few decoder layers are fused, the saliency maps may focus on partial object regions or contain scattered responses from surrounding context. As $L'$ increases, especially when more high-level and intermediate decoder layers are jointly used, the highlighted regions become better aligned with the target instances and less distracted by background activations. This indicates that the tracked query captures complementary attribution cues across different decoding stages, while the reliability-weighted fusion helps integrate these cues into a more stable instance-level explanation. 

\subsubsection{Effect of the Hybrid Fusion Strategy}
\label{sec:ablation_clhcf_fusion}
Table~\ref{tab:clhcf_ablation} compares different fusion variants of CLHCF. It can be observed that Geom., which uses only the geometric branch, achieves better Energy PG and Comp. than Mean and Arith. on all three datasets. This indicates that geometric fusion can better emphasize stable responses that repeatedly appear across layers while suppressing scattered activations that occur only in a few layers, thereby improving the concentration of saliency energy within the target region. In contrast, Arith., which uses only the arithmetic branch, does not bring consistent improvements. Its Energy PG is even lower than that of the simple inter-layer average Mean, and its Comp. is higher. This suggests that although arithmetic fusion can preserve complementary responses from different layers, it may also retain scattered activations or responses weakly related to the target when no consistency constraint is imposed.

On this basis, Hybrid combines the arithmetic and geometric branches and achieves the best results on all datasets. Compared with the strongest non-hybrid variant, Geom., Hybrid improves Energy PG by 13.75\%, 11.18\%, and 28.42\% on COCO, ExDark, and Cityscapes, respectively, while reducing Comp. by 29.97\%, 20.25\%, and 35.41\%. These results demonstrate that relying solely on geometric fusion can improve spatial concentration but may weaken some complementary cross-layer evidence. In contrast, Hybrid preserves target-related complementary responses through the arithmetic branch and suppresses unstable activations through the geometric branch, thereby forming a more effective hybrid consensus fusion strategy and generating instance-level explanation maps with more concentrated energy and more compact spatial distributions.

Fig.~\ref{fig:ablation_hybrid_visualization} provides a qualitative comparison among different CLHCF fusion variants. Mean and Arith. tend to preserve more activations across layers, but they also retain scattered responses outside the target object regions. Geom. produces more concentrated maps by leveraging cross-layer consensus, which effectively suppresses many layer-inconsistent responses. However, some residual scattered responses are still not fully removed. By contrast, Hybrid further attenuates these residual responses and reduces their spatial spread, while preserving the main target-related regions.

\section{Conclusion}
In this paper, we propose an End-to-end Instance-specific Visual Explanation framework (EIVE) for Detection Transformers. The proposed method models the cross-attention mechanism in the decoder as an instance-level feature attribution pathway, enabling the model to derive corresponding instance-level saliency maps from the forward-pass outputs after detection, without requiring additional gradient computation or input perturbation. As a result, the explanation efficiency is significantly improved. Building upon this idea, we further design a cross-layer hybrid consensus fusion (CLHCF) module to aggregate explanation information across decoder layers, thereby producing more stable, compact, and semantically consistent instance-level explanations. In addition, as a training-oriented application, we propose an attention-aware joint training strategy (AAJTS), which constrains the spatial distribution of cross-attention and guides the model to learn more focused instance-level attribution representations, leading to more faithful and compact explanations while further improving detection performance. Experimental results on the COCO, ExDark, and Cityscapes datasets demonstrate that the proposed method achieves strong performance in terms of explanation faithfulness, spatial localization ability, saliency map compactness, and computational efficiency. Moreover, its effectiveness and generality are validated on multiple DETR-like detection frameworks.
﻿

\bibliographystyle{IEEEtran}
\bibliography{EIVE}

@IEEEtranBSTCTL{BSTcontrol,
	CTLdash_repeated_names = "no"
}

@inproceedings{ref_Residual,
	title={Deep residual learning for image recognition},
	author={He, Kaiming and Zhang, Xiangyu and Ren, Shaoqing and Sun, Jian},
	booktitle={Proceedings of the IEEE conference on computer vision and pattern recognition},
	pages={770--778},
	year={2016}
}

@inproceedings{ref_Shufflenet,
	title={Shufflenet: An extremely efficient convolutional neural network for mobile devices},
	author={Zhang, Xiangyu and Zhou, Xinyu and Lin, Mengxiao and Sun, Jian},
	booktitle={Proceedings of the IEEE conference on computer vision and pattern recognition},
	pages={6848--6856},
	year={2018}
}

@inproceedings{ref_Mobilenetv2,
	title={Mobilenetv2: Inverted residuals and linear bottlenecks},
	author={Sandler, Mark and Howard, Andrew and Zhu, Menglong and Zhmoginov, Andrey and Chen, Liang-Chieh},
	booktitle={Proceedings of the IEEE conference on computer vision and pattern recognition},
	pages={4510--4520},
	year={2018}
}

@article{ref_SBSNet,
	title={SBSNet: Spatial-Spectral Background-Target Separation Network for Hyperspectral Target Detection},
	author={Xiang, Jianlin and Li, Yanshan and Dai, Linhui and Qi, Ruo and Tang, Haojin and Zhang, Li and Zhang, Kunhua and Xie, Weixin},
	journal={IEEE Journal of Selected Topics in Applied Earth Observations and Remote Sensing},
	year={2026},
	publisher={IEEE}
}

@InProceedings{ref_PRFA,
	author={Liang, Siyuan and Wu, Baoyuan and Fan, Yanbo and Wei, Xingxing and Cao, Xiaochun},
	title={Parallel Rectangle Flip Attack: A Query-Based Black-Box Attack Against Object Detection},
	booktitle={Proceedings of the IEEE/CVF International Conference on Computer Vision (ICCV)},
	month={October},
	year={2021},
	pages={7697-7707}
}

@inproceedings{ref_BBAAOB,
	title={A large-scale multiple-objective method for black-box attack against object detection},
	author={Liang, Siyuan and Li, Longkang and Fan, Yanbo and Jia, Xiaojun and Li, Jingzhi and Wu, Baoyuan and Cao, Xiaochun},
	booktitle={European Conference on Computer Vision},
	pages={619--636},
	year={2022},
	organization={Springer}
}

@article{ref_PODAD,
	title={A review and comparative study on probabilistic object detection in autonomous driving},
	author={Feng, Di and Harakeh, Ali and Waslander, Steven L and Dietmayer, Klaus},
	journal={IEEE Transactions on Intelligent Transportation Systems},
	volume={23},
	number={8},
	pages={9961--9980},
	year={2021},
	publisher={IEEE}
}

@inproceedings{ref_Safe,
	title={Safe: Sensitivity-aware features for out-of-distribution object detection},
	author={Wilson, Samuel and Fischer, Tobias and Dayoub, Feras and Miller, Dimity and S{\"u}nderhauf, Niko},
	booktitle={Proceedings of the ieee/cvf international conference on computer vision},
	pages={23565--23576},
	year={2023}
}

@article{ref_GT-CAM,
	title={Gt-cam: Game theory based class activation map for gcn},
	author={Li, Yanshan and Shi, Ting and Chen, Zhiyuan and Zhang, Li and Xie, Weixin},
	journal={IEEE Transactions on Pattern Analysis and Machine Intelligence},
	volume={46},
	number={12},
	pages={8806--8819},
	year={2024},
	publisher={IEEE}
}

@article{ref_BI-CAM,
	title={BI-CAM: Generating explanations for deep neural networks using bipolar information},
	author={Li, Yanshan and Liang, Huajie and Yu, Rui},
	journal={IEEE Transactions on Multimedia},
	volume={26},
	pages={568--580},
	year={2023},
	publisher={IEEE}
}

@article{ref_CR-CAM,
	title={CR-CAM: Generating explanations for deep neural networks by contrasting and ranking features},
	author={Li, Yanshan and Liang, Huajie and Zheng, Hongfang and Yu, Rui},
	journal={Pattern Recognition},
	volume={149},
	pages={110251},
	year={2024},
	publisher={Elsevier}
}

@inproceedings{ref_CAM,
	title={Learning deep features for discriminative localization},
	author={Zhou, Bolei and Khosla, Aditya and Lapedriza, Agata and Oliva, Aude and Torralba, Antonio},
	booktitle={Proceedings of the IEEE conference on computer vision and pattern recognition},
	pages={2921--2929},
	year={2016}
}

@article{ref_LayerCAM,
	title={Layercam: Exploring hierarchical class activation maps for localization},
	author={Jiang, Peng-Tao and Zhang, Chang-Bin and Hou, Qibin and Cheng, Ming-Ming and Wei, Yunchao},
	journal={IEEE transactions on image processing},
	volume={30},
	pages={5875--5888},
	year={2021},
	publisher={IEEE}
}

@inproceedings{ref_Grad-CAM,
	title={Grad-cam: Visual explanations from deep networks via gradient-based localization},
	author={Selvaraju, Ramprasaath R and Cogswell, Michael and Das, Abhishek and Vedantam, Ramakrishna and Parikh, Devi and Batra, Dhruv},
	booktitle={Proceedings of the IEEE international conference on computer vision},
	pages={618--626},
	year={2017}
}

@inproceedings{ref_Grad-CAM++,
	title={Grad-cam++: Generalized gradient-based visual explanations for deep convolutional networks},
	author={Chattopadhay, Aditya and Sarkar, Anirban and Howlader, Prantik and Balasubramanian, Vineeth N},
	booktitle={2018 IEEE winter conference on applications of computer vision (WACV)},
	pages={839--847},
	year={2018},
	organization={IEEE}
}

@inproceedings{ref_Score-CAM,
	title={Score-CAM: Score-weighted visual explanations for convolutional neural networks},
	author={Wang, Haofan and Wang, Zifan and Du, Mengnan and Yang, Fan and Zhang, Zijian and Ding, Sirui and Mardziel, Piotr and Hu, Xia},
	booktitle={Proceedings of the IEEE/CVF conference on computer vision and pattern recognition workshops},
	pages={24--25},
	year={2020}
}

@inproceedings{ref_SSGrad-CAM++,
	title={Spatial Sensitive Grad-CAM++: Improved visual explanation for object detectors via weighted combination of gradient map},
	author={Yamauchi, Toshinori},
	booktitle={Proceedings of the IEEE/CVF Conference on Computer Vision and Pattern Recognition},
	pages={8164--8168},
	year={2024}
}

@inproceedings{ref_Finer-CAM,
	title={Finer-cam: Spotting the difference reveals finer details for visual explanation},
	author={Zhang, Ziheng and Gu, Jianyang and Chowdhury, Arpita and Mai, Zheda and Carlyn, David and Berger-Wolf, Tanya and Su, Yu and Chao, Wei-Lun},
	booktitle={Proceedings of the Computer Vision and Pattern Recognition Conference},
	pages={9611--9620},
	year={2025}
}

@inproceedings{ref_LIME,
	title={" Why should i trust you?" Explaining the predictions of any classifier},
	author={Ribeiro, Marco Tulio and Singh, Sameer and Guestrin, Carlos},
	booktitle={Proceedings of the 22nd ACM SIGKDD international conference on knowledge discovery and data mining},
	pages={1135--1144},
	year={2016}
}

@article{ref_RISE,
	title={Rise: Randomized input sampling for explanation of black-box models},
	author={Petsiuk, Vitali and Das, Abir and Saenko, Kate},
	journal={arXiv preprint arXiv:1806.07421},
	year={2018}
}

@inproceedings{fong2017interpretable,
	title={Interpretable explanations of black boxes by meaningful perturbation},
	author={Fong, Ruth C and Vedaldi, Andrea},
	booktitle={Proceedings of the IEEE international conference on computer vision},
	pages={3429--3437},
	year={2017}
}

@inproceedings{fong2019understanding,
	title={Understanding deep networks via extremal perturbations and smooth masks},
	author={Fong, Ruth and Patrick, Mandela and Vedaldi, Andrea},
	booktitle={Proceedings of the IEEE/CVF international conference on computer vision},
	pages={2950--2958},
	year={2019}
}

@inproceedings{ref_IGOS,
	title={Visualizing Deep Networks by Optimizing with Integrated Gradients.},
	author={Qi, Zhongang and Khorram, Saeed and Li, Fuxin},
	booktitle={AAAI},
	volume={34},
	pages={11890--11898},
	year={2020}
}

@inproceedings{ref_IGOS++,
	title={iGOS++ integrated gradient optimized saliency by bilateral perturbations},
	author={Khorram, Saeed and Lawson, Tyler and Fuxin, Li},
	booktitle={Proceedings of the Conference on Health, Inference, and Learning},
	pages={174--182},
	year={2021}
}

@article{ref_SHAP,
	title={A unified approach to interpreting model predictions},
	author={Lundberg, Scott M and Lee, Su-In},
	journal={Advances in neural information processing systems},
	volume={30},
	year={2017}
}

@article{ref_DeepSHAP,
	title={Consistent individualized feature attribution for tree ensembles},
	author={Lundberg, Scott M and Erion, Gabriel G and Lee, Su-In},
	journal={arXiv preprint arXiv:1802.03888},
	year={2018}
}

@inproceedings{ref_RCNN,
	title={Rich feature hierarchies for accurate object detection and semantic segmentation},
	author={Girshick, Ross and Donahue, Jeff and Darrell, Trevor and Malik, Jitendra},
	booktitle={Proceedings of the IEEE conference on computer vision and pattern recognition},
	pages={580--587},
	year={2014}
}

@inproceedings{ref_FastRCNN,
	title={Fast r-cnn},
	author={Girshick, Ross},
	booktitle={Proceedings of the IEEE international conference on computer vision},
	pages={1440--1448},
	year={2015}
}

@article{ref_FasterRCNN,
	title={Faster R-CNN: Towards real-time object detection with region proposal networks},
	author={Ren, Shaoqing and He, Kaiming and Girshick, Ross and Sun, Jian},
	journal={IEEE transactions on pattern analysis and machine intelligence},
	volume={39},
	number={6},
	pages={1137--1149},
	year={2016},
	publisher={IEEE}
}

@inproceedings{ref_Mask_RCNN,
	title={Mask r-cnn},
	author={He, Kaiming and Gkioxari, Georgia and Doll{\'a}r, Piotr and Girshick, Ross},
	booktitle={Proceedings of the IEEE international conference on computer vision},
	pages={2961--2969},
	year={2017}
}

@inproceedings{ref_YOLO,
	title={You only look once: Unified, real-time object detection},
	author={Redmon, Joseph and Divvala, Santosh and Girshick, Ross and Farhadi, Ali},
	booktitle={Proceedings of the IEEE conference on computer vision and pattern recognition},
	pages={779--788},
	year={2016}
}

@inproceedings{ref_RetinaNet,
	title={Focal loss for dense object detection},
	author={Lin, Tsung-Yi and Goyal, Priya and Girshick, Ross and He, Kaiming and Doll{\'a}r, Piotr},
	booktitle={Proceedings of the IEEE international conference on computer vision},
	pages={2980--2988},
	year={2017}
}

@inproceedings{ref_FCOS,
	title={Fcos: Fully convolutional one-stage object detection},
	author={Tian, Zhi and Shen, Chunhua and Chen, Hao and He, Tong},
	booktitle={Proceedings of the IEEE/CVF international conference on computer vision},
	pages={9627--9636},
	year={2019}
}

@inproceedings{ref_D-RISE,
	title={Black-box explanation of object detectors via saliency maps},
	author={Petsiuk, Vitali and Jain, Rajiv and Manjunatha, Varun and Morariu, Vlad I and Mehra, Ashutosh and Ordonez, Vicente and Saenko, Kate},
	booktitle={Proceedings of the IEEE/CVF conference on computer vision and pattern recognition},
	pages={11443--11452},
	year={2021}
}

@article{ref_D-HSIC,
	title={Making sense of dependence: Efficient black-box explanations using dependence measure},
	author={Novello, Paul and Fel, Thomas and Vigouroux, David},
	journal={Advances in Neural Information Processing Systems},
	volume={35},
	pages={4344--4357},
	year={2022}
}

@inproceedings{ref_VPS,
	title={Interpreting object-level foundation models via visual precision search},
	author={Chen, Ruoyu and Liang, Siyuan and Li, Jingzhi and Liu, Shiming and Li, Maosen and Huang, Zhen and Zhang, Hua and Cao, Xiaochun},
	booktitle={Proceedings of the Computer Vision and Pattern Recognition Conference},
	pages={30042--30052},
	year={2025}
}

@article{ref_Transformer,
	title={Attention is all you need},
	author={Vaswani, Ashish and Shazeer, Noam and Parmar, Niki and Uszkoreit, Jakob and Jones, Llion and Gomez, Aidan N and Kaiser, {\L}ukasz and Polosukhin, Illia},
	journal={Advances in neural information processing systems},
	volume={30},
	year={2017}
}

@inproceedings{jain2019attention,
	title={Attention is not explanation},
	author={Jain, Sarthak and Wallace, Byron C},
	booktitle={Proceedings of the 2019 Conference of the North American Chapter of the Association for Computational Linguistics: Human Language Technologies, Volume 1 (Long and Short Papers)},
	pages={3543--3556},
	year={2019}
}

@article{ref_ViT,
	title={An image is worth 16x16 words: Transformers for image recognition at scale},
	author={Dosovitskiy, Alexey and Beyer, Lucas and Kolesnikov, Alexander and Weissenborn, Dirk and Zhai, Xiaohua and Unterthiner, Thomas and Dehghani, Mostafa and Minderer, Matthias and Heigold, Georg and Gelly, Sylvain and others},
	journal={arXiv preprint arXiv:2010.11929},
	year={2020}
}

@inproceedings{ref_DeiT,
	title={Training data-efficient image transformers \& distillation through attention},
	author={Touvron, Hugo and Cord, Matthieu and Douze, Matthijs and Massa, Francisco and Sablayrolles, Alexandre and J{\'e}gou, Herv{\'e}},
	booktitle={International conference on machine learning},
	pages={10347--10357},
	year={2021},
	organization={PMLR}
}

@inproceedings{ref_Swin,
	title={Swin transformer: Hierarchical vision transformer using shifted windows},
	author={Liu, Ze and Lin, Yutong and Cao, Yue and Hu, Han and Wei, Yixuan and Zhang, Zheng and Lin, Stephen and Guo, Baining},
	booktitle={Proceedings of the IEEE/CVF international conference on computer vision},
	pages={10012--10022},
	year={2021}
}

@inproceedings{ref_PVT,
	title={Pyramid vision transformer: A versatile backbone for dense prediction without convolutions},
	author={Wang, Wenhai and Xie, Enze and Li, Xiang and Fan, Deng-Ping and Song, Kaitao and Liang, Ding and Lu, Tong and Luo, Ping and Shao, Ling},
	booktitle={Proceedings of the IEEE/CVF international conference on computer vision},
	pages={568--578},
	year={2021}
}

@inproceedings{ref_CvT,
	title={Cvt: Introducing convolutions to vision transformers},
	author={Wu, Haiping and Xiao, Bin and Codella, Noel and Liu, Mengchen and Dai, Xiyang and Yuan, Lu and Zhang, Lei},
	booktitle={Proceedings of the IEEE/CVF international conference on computer vision},
	pages={22--31},
	year={2021}
}

@InProceedings{Chefer_2021_ICCV,
	title     = {Generic Attention-Model Explainability for Interpreting Bi-Modal and Encoder-Decoder Transformers},
	author    = {Chefer, Hila and Gur, Shir and Wolf, Lior},
	booktitle = {Proceedings of the IEEE/CVF International Conference on Computer Vision (ICCV)},
	month     = {October},
	year      = {2021},
	pages     = {397-406}
}

@article{ref_AttCAT,
	title={Attcat: Explaining transformers via attentive class activation tokens},
	author={Qiang, Yao and Pan, Deng and Li, Chengyin and Li, Xin and Jang, Rhongho and Zhu, Dongxiao},
	journal={Advances in neural information processing systems},
	volume={35},
	pages={5052--5064},
	year={2022}
}

@inproceedings{ref_TokenTM,
	title={Token transformation matters: Towards faithful post-hoc explanation for vision transformer},
	author={Wu, Junyi and Duan, Bin and Kang, Weitai and Tang, Hao and Yan, Yan},
	booktitle={Proceedings of the IEEE/CVF Conference on Computer Vision and Pattern Recognition},
	pages={10926--10935},
	year={2024}
}

@inproceedings{ref_FaithfulnessViT,
	title={On the faithfulness of vision transformer explanations},
	author={Wu, Junyi and Kang, Weitai and Tang, Hao and Hong, Yuan and Yan, Yan},
	booktitle={Proceedings of the IEEE/CVF Conference on Computer Vision and Pattern Recognition},
	pages={10936--10945},
	year={2024}
}

@inproceedings{ref_DETR,
	title={End-to-end object detection with transformers},
	author={Carion, Nicolas and Massa, Francisco and Synnaeve, Gabriel and Usunier, Nicolas and Kirillov, Alexander and Zagoruyko, Sergey},
	booktitle={European conference on computer vision},
	pages={213--229},
	year={2020},
	organization={Springer}
}

@article{ref_deformable,
	title={Deformable detr: Deformable transformers for end-to-end object detection},
	author={Zhu, Xizhou and Su, Weijie and Lu, Lewei and Li, Bin and Wang, Xiaogang and Dai, Jifeng},
	journal={arXiv preprint arXiv:2010.04159},
	year={2020}
}

@inproceedings{ref_conditional,
	title={Conditional detr for fast training convergence},
	author={Meng, Depu and Chen, Xiaokang and Fan, Zejia and Zeng, Gang and Li, Houqiang and Yuan, Yuhui and Sun, Lei and Wang, Jingdong},
	booktitle={Proceedings of the IEEE/CVF international conference on computer vision},
	pages={3651--3660},
	year={2021}
}

@article{ref_dab,
	title={Dab-detr: Dynamic anchor boxes are better queries for detr},
	author={Liu, Shilong and Li, Feng and Zhang, Hao and Yang, Xiao and Qi, Xianbiao and Su, Hang and Zhu, Jun and Zhang, Lei},
	journal={arXiv preprint arXiv:2201.12329},
	year={2022}
}

@article{ref_dino,
	title={Dino: Detr with improved denoising anchor boxes for end-to-end object detection},
	author={Zhang, Hao and Li, Feng and Liu, Shilong and Zhang, Lei and Su, Hang and Zhu, Jun and Ni, Lionel M and Shum, Heung-Yeung},
	journal={arXiv preprint arXiv:2203.03605},
	year={2022}
}

@inproceedings{ref_Attention-Rollout,
	title={Quantifying attention flow in transformers},
	author={Abnar, Samira and Zuidema, Willem},
	booktitle={Proceedings of the 58th annual meeting of the association for computational linguistics},
	pages={4190--4197},
	year={2020}
}

@inproceedings{ref_Attention-Visualization,
	title={Transformer interpretability beyond attention visualization},
	author={Chefer, Hila and Gur, Shir and Wolf, Lior},
	booktitle={Proceedings of the IEEE/CVF conference on computer vision and pattern recognition},
	pages={782--791},
	year={2021}
}

@article{ref_DETD,
	title={Explainability enhanced object detection transformer with feature disentanglement},
	author={Yu, Wenlong and Liu, Ruonan and Chen, Dongyue and Hu, Qinghua},
	journal={IEEE Transactions on Image Processing},
	volume={33},
	pages={6439--6454},
	year={2024},
	publisher={IEEE}
}

@article{ref_PG,
	title={Top-down neural attention by excitation backprop},
	author={Zhang, Jianming and Bargal, Sarah Adel and Lin, Zhe and Brandt, Jonathan and Shen, Xiaohui and Sclaroff, Stan},
	journal={International Journal of Computer Vision},
	volume={126},
	number={10},
	pages={1084--1102},
	year={2018},
	publisher={Springer}
}

@article{ref_ODAM,
	title={Gradient-based instance-specific visual explanations for object specification and object discrimination},
	author={Zhao, Chenyang and Hsiao, Janet H and Chan, Antoni B},
	journal={IEEE Transactions on Pattern Analysis and Machine Intelligence},
	volume={46},
	number={9},
	pages={5967--5985},
	year={2024},
	publisher={IEEE}
}

@inproceedings{ref_COCO,
	title={Microsoft coco: Common objects in context},
	author={Lin, Tsung-Yi and Maire, Michael and Belongie, Serge and Hays, James and Perona, Pietro and Ramanan, Deva and Doll{\'a}r, Piotr and Zitnick, C Lawrence},
	booktitle={European conference on computer vision},
	pages={740--755},
	year={2014},
	organization={Springer}
}

@article{ref_Exdark,
	title={Getting to know low-light images with the exclusively dark dataset},
	author={Loh, Yuen Peng and Chan, Chee Seng},
	journal={Computer vision and image understanding},
	volume={178},
	pages={30--42},
	year={2019},
	publisher={Elsevier}
}

@inproceedings{ref_Cityscapes,
	title={The cityscapes dataset for semantic urban scene understanding},
	author={Cordts, Marius and Omran, Mohamed and Ramos, Sebastian and Rehfeld, Timo and Enzweiler, Markus and Benenson, Rodrigo and Franke, Uwe and Roth, Stefan and Schiele, Bernt},
	booktitle={Proceedings of the IEEE conference on computer vision and pattern recognition},
	pages={3213--3223},
	year={2016}
}

\end{document}